\documentclass[11pt]{article}

\usepackage[preprint]{acl}

\usepackage{times}
\usepackage{latexsym}
\usepackage[T1]{fontenc}
\usepackage[utf8]{inputenc}
\usepackage{microtype}
\usepackage{inconsolata}
\usepackage{graphicx}
\usepackage{booktabs}
\usepackage{multirow}
\usepackage{colortbl}
\usepackage{amsmath}
\usepackage{amssymb}
\usepackage{algorithm}
\usepackage{algpseudocode}

\newcommand{\methodname}{ReWEIGH}
\newcommand{\llavaonefive}{LLaVA-1.5-7B}
\newcommand{\qwenvlseven}{Qwen2.5-VL-7B}
\newcommand{\instructblipseven}{InstructBLIP-7B}
\newcommand{\llavanextseven}{LLaVA-NeXT-7B}
\newcommand{\dmrr}{\operatorname{DMRR}}
\newcommand{\rank}{\operatorname{rank}}
\newcommand{\Norm}{\operatorname{Norm}}
\newcommand{\Cand}{\mathcal{C}}
\newcommand{\clip}{\operatorname{clip}}
\newcommand{\vocab}{\mathcal{V}}

\title{\methodname{} the Evidence: Calibrating Token-Level Ordinal Visual Evidence to Mitigate Hallucinations in Large Vision-Language Models}

\author{
    Jihae Jeong\Thanks{Both authors contributed equally to this work.} 
    \quad 
    Junha Choi\footnotemark[1] 
    \quad  
    Hwanjo Yu\Thanks{Corresponding author.} \\
    Pohang University of Science and Technology (POSTECH) \quad \\
    \texttt{\{wisdomjeong, 2002junha, hwanjoyu\}@postech.ac.kr}
}

\begin{document}
\maketitle

\begin{abstract}

Large vision-language models (LVLMs) often hallucinate, generating content that the input image does not support.
Preventing such content during decoding calls for a candidate-specific measure of how strongly the image supports the token under consideration.
The model's visual-token states offer a natural source of this evidence because projecting each state through the output head reveals which vocabulary items that position favors.
These position-wise readouts cannot be pooled directly because their probability magnitudes are not comparable across visual positions.
Vocabulary ranks provide a scale-invariant basis for pooling, but tokens still differ systematically in their typical rank-based evidence.
We propose \methodname{}, a training-free decoding intervention that aggregates these ranks across visual positions and compares each candidate with a token-specific reference estimated from unlabeled images.
At inference, \methodname{} caches the image evidence during prefill and applies a bounded penalty only to candidates that fall below their reference.
On four 7B backbones, \methodname{} reduces hallucinated object mentions by up to $21.3\%$ while largely preserving or improving descriptive and general performance.
With evidence cached, the average added latency is $1.33\%$ per token, and the reductions extend across six architecture families to 32B parameters.

\end{abstract}

\section{Introduction}
\label{sec:introduction}

Large vision-language models (LVLMs) have emerged as general-purpose systems for visual understanding and reasoning \citep{liu2023llava,dai2023instructblip}.
However, they often hallucinate content that conflicts with the input image or lacks visual support \citep{li2023pope,wang2023amber}.
These errors can arise when strong linguistic priors outweigh evidence from the image \citep{leng2024vcd,favero2024m3id}.
Because the resulting text remains fluent and plausible, the text alone often provides little indication of the error.
This failure mode motivates interventions that act during decoding, before the model emits content without visual support.

Existing decoding interventions face a trade-off between computational cost and grounding specificity.
Contrastive methods compute corrective logits under altered visual or textual conditioning, which commonly requires additional forward or decoding passes \citep{leng2024vcd,favero2024m3id,wang2024icd,kim2024code}.
Lighter methods use attention to control reliance on visual tokens \citep{huang2024opera,liu2024pai} or use uncertainty to decide when to reinject visual features \citep{zou2025memvr}.
These methods assume that attention or uncertainty indicates insufficient grounding.
Attention reveals where the model routes information, while uncertainty reflects confidence in the output distribution.
Neither signal directly measures how strongly the image supports a particular candidate token.
Output confidence also misses many errors, as hallucinated object mentions persist even in its top quartile (Table~\ref{tab:confidence-blindness}, Appendix~\ref{sec:additional-analyses}).
Existing methods therefore offer no lightweight, candidate-specific measure of whether the model's visual states provide unusually weak support for the token under consideration.

The LVLM's internal visual states provide a natural source of token-specific evidence without auxiliary model passes.
Applying the output head to each visual-token state yields a vocabulary readout that reveals which words each visual position favors \citep{nostalgebraist2020logitlens,jiang2025interpreting}.
Prior lens-guided decoding uses the resulting probability magnitudes \citep{cho-kim-2026-revisit}, but their values depend on the sharpness of each position's output distribution.
Pooling these probabilities can therefore give disproportionate weight to sharper visual positions.
Vocabulary ranks avoid this scale dependence and provide a stable ordinal basis for aggregation.
However, ranks do not support a single criterion for all tokens.
Some tokens routinely rank near the top across visual positions, while others have lower typical evidence.
Visual support must therefore be assessed against the candidate token's own evidence baseline.

Based on this analysis, we propose \methodname{} (\textbf{Re}ference-calibrated \textbf{W}eighting via Ordinal \textbf{E}vidence for \textbf{I}mage-\textbf{G}rounded \textbf{H}allucination Mitigation), a training-free decoding intervention.
During offline calibration on unlabeled images, \methodname{} aggregates the visual readouts using dense mean reciprocal rank (DMRR) and estimates each candidate token's typical evidence level.
It registers a token for intervention only when this reference estimate is stable.
At inference, \methodname{} computes and caches the image's DMRR evidence during prefill.
When a registered candidate falls below its reference, the method applies a bounded logit penalty.

Across four 7B backbones, \methodname{} reduces CHAIR$_I$ \citep{rohrbach2018object} by up to $21.3\%$ while largely preserving or improving descriptive and general multimodal performance.
The reductions extend to 11 models from six architecture families, with model sizes ranging from 7B to 32B.
Each model requires a one-time offline calibration on unlabeled images.
With image-level evidence cached, the average added latency is $1.33\%$ per token.

Our contributions are as follows:
\begin{itemize}
    \item We show that probability pooling remains sensitive to position-specific distribution shapes after affine scale equalization. Rank aggregation is invariant to strictly increasing transformations at each position and provides a stable basis for pooling visual readouts.
    \item We identify a strong token-dependent component in ordinal visual evidence. We then introduce candidate-conditioned, token-specific calibration that assesses each token against its typical evidence baseline.
    \item We develop a bounded, training-free decoding intervention that reduces hallucination across diverse LVLM backbones and architectures while largely preserving descriptive and general multimodal utility with minimal inference overhead.
\end{itemize}

\section{Related Work}
\label{sec:related-work}

Training-free decoding methods mitigate LVLM hallucinations without updating the backbone.
Contrastive methods derive reference logits from altered visual inputs or views \citep{leng2024vcd,favero2024m3id,chen2024halc}, altered instructions or self-generated descriptions \citep{wang2024icd,kim2024code}, or selectively retained vision tokens \citep{huo2025sid}.
These approaches typically require an auxiliary decoding trajectory and make the corrective signal depend on how the counterfactual condition is constructed.
Other methods control attention or visual-token reliance \citep{huang2024opera,liu2024pai,gong2024damro} or use uncertainty-triggered reinjection of visual features \citep{zou2025memvr}.
Although these signals can strengthen visual conditioning, they do not directly provide a token-specific reference for determining whether a candidate has unusually weak support from the image.

Our approach belongs most directly to internal-representation interventions.
Vocabulary projection maps hidden states into language space through the output head \citep{nostalgebraist2020logitlens} and, in LVLMs, reveals signals that help distinguish visually grounded objects from hallucinated ones \citep{jiang2025interpreting}.
ReVisiT selects one vision token at each decoding step and uses its constrained probability projection to refine the output \citep{cho-kim-2026-revisit}.
Consequently, evidence distributed across the remaining visual positions is not explicitly aggregated.
DeCo instead blends earlier-layer predictions from the current generation position \citep{wang2025deco}, rather than directly summarizing visual-token states.
Activation Steering Decoding learns hallucination directions from labeled hidden states and contrasts predictions from opposite steering passes \citep{su2025activation}, but does not measure candidate-specific support from the image.
In contrast, we aggregate scale-invariant ordinal readouts over all visual positions once during prefill, cache the resulting image-level evidence throughout generation, and interpret each candidate relative to its token-specific reference estimated from unlabeled calibration data.

\section{Two Obstacles to Decoding with Internal Visual Evidence}

\label{sec:motivation}

\begin{figure}[t]
    \centering
    \includegraphics[width=0.93\columnwidth]{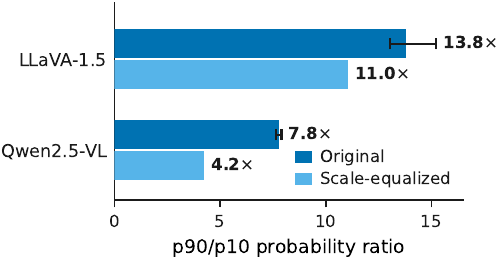}
    \caption{\textbf{Across-position spread of the rank-10 probability
    before and after scale equalization.} Error bars are retrospective
    image-bootstrap 95\% intervals, and equalized bars are point estimates.}
    \label{fig:scale-nuisance}
\end{figure}

\begin{figure}[t]
    \centering
    \includegraphics[width=0.93\columnwidth]{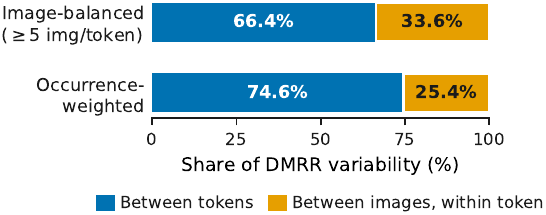}
    \caption{\textbf{Decomposition of DMRR variability on \llavaonefive{}.}
    Each bar sums to 100\%.}
    \label{fig:token-decomposition}
\end{figure}

\paragraph{Visual Evidence from Internal States.}
LVLMs frequently hallucinate unsupported content in open-ended descriptions \citep{li2023pope,wang2023amber}, but their internal visual states can provide a proxy for the visual evidence associated with individual words.
Projecting a visual-position hidden state through the language model's output head yields a vocabulary-space readout of the words associated with that position \citep{nostalgebraist2020logitlens}.
Such readouts distinguish grounded objects from hallucinated ones \citep{jiang2025interpreting}, and grounded alternatives remain highly ranked even at decoding steps that produce hallucinations \citep{cho-kim-2026-revisit}.

Formally, let $P$ denote the set of visual-token positions and $\mathbf h_j^{(\ell)}$ the hidden state at position $j\in P$ and language-model layer $\ell$.
Following the logit-lens view, we apply the model's final output normalization $\Norm$ and vocabulary projection $\mathbf W_{\mathrm{head}}$ to define the visual-position readout
\begin{equation}
    \mathbf z_j^{(\ell)}
    = \mathbf W_{\mathrm{head}}\!\left(
        \Norm\!\left(\mathbf h_j^{(\ell)}\right)
      \right).
    \label{eq:lens}
\end{equation}
Each component of $\mathbf z_j^{(\ell)}$ is the vocabulary score that visual position $j$ assigns to the corresponding vocabulary item.

Turning these internal readouts into a decoding intervention, however, requires two steps.
First, we must aggregate evidence from many visual positions without distortion (Section~\ref{sec:scale}).
Second, we must interpret the aggregated value relative to each candidate token (Section~\ref{sec:offset}).
These two problems determine \methodname{}'s evidence measurement and per-candidate calibration, respectively.

\subsection{Scale-Invariant Aggregation Across Visual Positions}
\label{sec:scale}

Probability magnitudes are not directly comparable across visual positions.
A direct approach applies softmax to Eq.~\eqref{eq:lens} and averages a token's probability across positions.
Prior lens-guided decoding likewise represents visual evidence using these probability magnitudes \citep{cho-kim-2026-revisit,jiang2025interpreting}.
However, the average inherits the sharpness of each position-wise distribution.
At a given position, a small probability can indicate weak support for the candidate.
It can also result from a distribution concentrated on another token.
To measure this dependence, we use 500 development images and record, for each image, the probability of the token at vocabulary rank 10 at every visual position.
The dispersion persists from rank 1 through rank 1,000 (Appendix~\ref{sec:rank-scale-analysis}).
Figure~\ref{fig:scale-nuisance} reports the image-median ratio between the 90th and 10th percentiles of this probability across positions.
The ratio is $13.8\times$ for \llavaonefive{} \citep{liu2024improved} and $7.8\times$ for \qwenvlseven{} \citep{bai2025qwen25vl}.
Thus, the probability at the same ordinal position can differ by up to an order of magnitude across visual positions.
Equalizing each position's logit scale to the image median reduces these ratios only to $11.0\times$ and $4.2\times$, respectively, leaving substantial dispersion in place. Therefore, an affine correction alone is insufficient.

This dependence propagates to the intervention itself.
We test sensitivity to position-specific scale while preserving vocabulary order.
We retain each position's readout direction but reassign the observed positive scale factors among visual positions.
We then reapply the same calibrated edit rule.
This rescaling does not change the vocabulary ordering within any position.
Any resulting change in the edit is therefore due to scale alone.
Section~\ref{sec:suppress} defines suppression strength as the per-candidate edit magnitude in $[0,1]$.
Under the probability readout, its mean absolute change is $0.045$ in \llavaonefive{} and $0.046$ in \qwenvlseven{}.
As shown in Table~\ref{tab:rescale-invariance} of Appendix~\ref{sec:rank-scale-analysis}, the rank-derived suppression strengths remain exactly unchanged.

Ranks, by contrast, are preserved under any strictly increasing transformation at each position.
Our narrow claim is that ordinal information provides a transformation-invariant basis for aggregating evidence across visual positions.
We pool visual evidence through reciprocal ranks, formalized in Section~\ref{sec:dmrr} as dense mean reciprocal rank (DMRR).
Among order-only statistics, DMRR is a simple parameter-free choice requiring no truncation depth or fitted weights.
It emphasizes positions that rank the token near the top while compressing differences among deep ranks.

\begin{figure}[t]
    \centering
    \includegraphics[width=0.93\columnwidth]{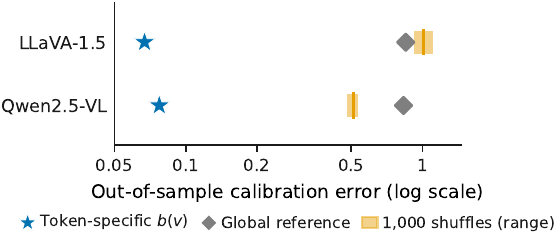}
    \caption{\textbf{Out-of-sample calibration error of token-specific,
    global, and shuffled references.} Errors are in $b_0$ units with every
    token counted once. Shading spans the full shuffle range, and the tick
    marks its median.}
    \label{fig:token-correspondence}
\end{figure}

\subsection{Token-Specific Interpretation of Aggregated Evidence}
\label{sec:offset}

Replacing magnitude with rank makes visual positions comparable, but it does not make vocabulary items interchangeable.
Some tokens routinely appear near the head of visual-position rankings.
Others have lower typical levels.
Figure~\ref{fig:token-decomposition} decomposes DMRR variability over calibration observations of \llavaonefive{}.
Each observation is an image-level DMRR value from a decoding step where the token is a plausible next-token candidate.
In the image-balanced decomposition, every pair of a token and an image contributes equally.
Differences between tokens explain $66.4\%$ of the variability, rather than differences between images for the same token.
A single global reference therefore conflates evidence that is unusually low for one token with evidence that is entirely typical for another.

This structure suggests using each token's calibration median as its reference.
To assess whether this token-level structure generalizes beyond the calibration data, we fit token-specific medians on 500 unlabeled MS~COCO images \citep{lin2014coco} and evaluate them on 4,969 disjoint images.
For each token, the calibration error is the gap between its held-out median evidence and the stored reference.
Relative to a single global reference, the token-specific medians reduce this error by $92\%$ in \llavaonefive{} and $91\%$ in \qwenvlseven{}.
They also outperform all 1,000 random reassignments of references to tokens (Figure~\ref{fig:token-correspondence}, Appendix~\ref{sec:token-reference-analysis}).
The gain comes from the correspondence between tokens and their references, not from the distribution of reference values alone.

References estimated from few or highly variable observations can misstate typical evidence and risk suppressing well-grounded candidates.
\methodname{} therefore registers a token only when uncertainty in its estimated reference has little effect on the resulting edit, abstaining otherwise (Section~\ref{sec:calib}).

Together, these diagnostics motivate ordinal aggregation, a token-specific reference, and a stability safeguard for uncertain references.
Section~\ref{sec:method} turns these ingredients into a bounded decoding intervention.

\section{\methodname: Token-Calibrated Ordinal Suppression}
\label{sec:method}

\subsection{Overview}
\label{sec:method-overview}

\begin{figure*}[t]
    \centering
    \includegraphics[width=0.96\textwidth]{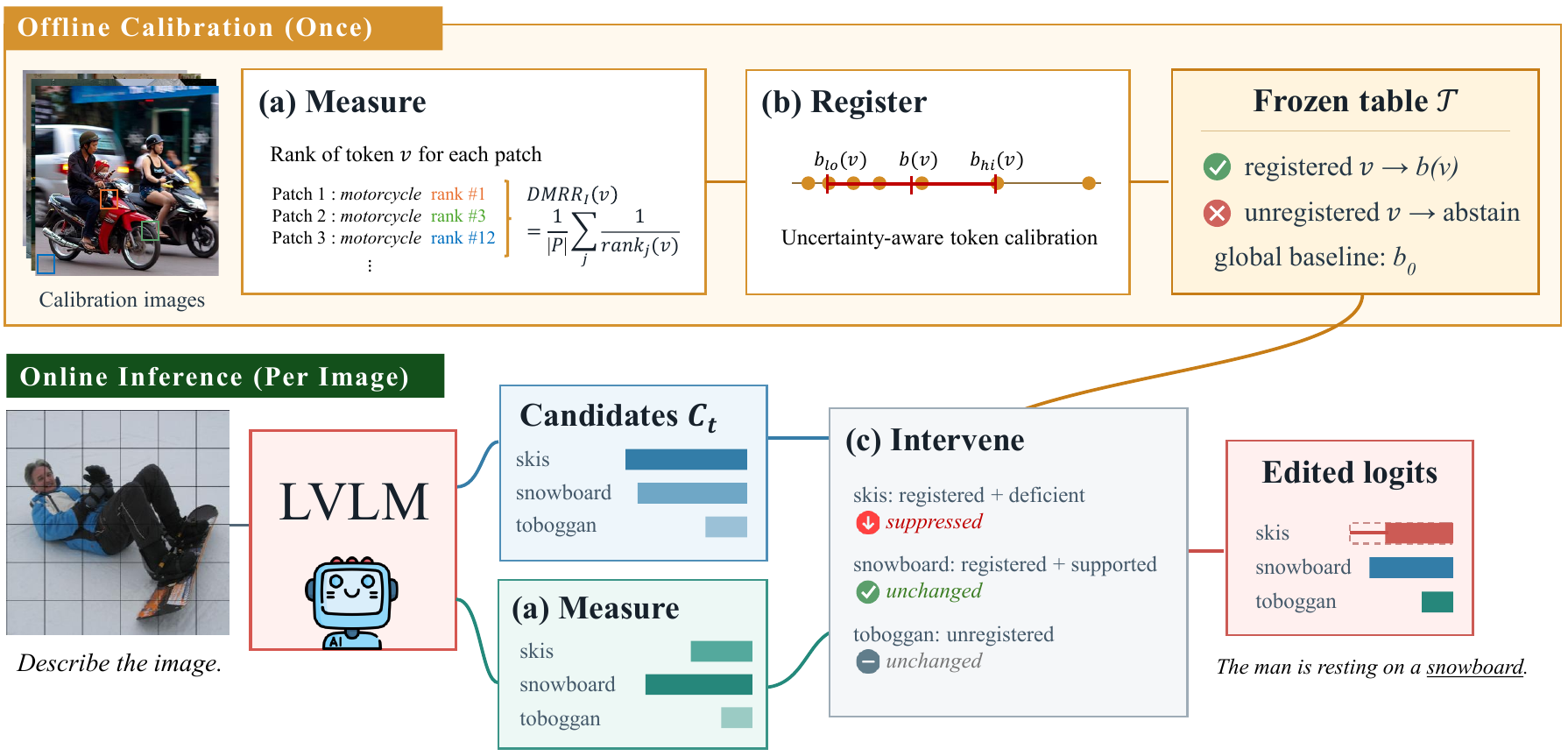}
    \caption{\textbf{Overview of \methodname{}.} \textbf{Measure} reads
    visual-token representations as vocabulary rankings. \textbf{Register}
    constructs a reliable token-specific reference table from unlabeled
    calibration images. \textbf{Intervene} applies a bounded negative shift
    to candidates with an evidence deficit. Measure is shared by offline
    calibration and online inference, whereas registration runs only once.}
    \label{fig:method}
\end{figure*}

Figure~\ref{fig:method} summarizes \methodname{}'s
\emph{Measure--Register--Intervene} workflow.
Offline, Measure aggregates visual-token readouts and Register builds a
reliability-filtered token-reference table from unlabeled images.
At inference, Measure computes image-level evidence once during prefill,
and Intervene uses the cached evidence and frozen table throughout decoding.
The resulting training-free method requires no additional model forward pass or external verifier.
Algorithm~\ref{alg:method} provides the complete procedure.

\subsection{Measure: Dense Ordinal Visual Evidence}
\label{sec:dmrr}

The Measure module quantifies how strongly the image-token states represent token $v$.
Let $\vocab$ denote the language model's vocabulary.
Recall from Eq.~\eqref{eq:lens} that $\mathbf z_j^{(\ell)}\in\mathbb R^{|\vocab|}$ is the LM-head readout of visual position $j\in P$ at layer $\ell$.
Let $\rank_j(v)$ denote the rank of $z_j^{(\ell)}(v)$ among the vocabulary scores in descending order, such that the largest score has rank 1.
We define the image-level evidence for token $v$ as its dense mean reciprocal rank (DMRR) across visual positions:
\begin{equation}
    \dmrr_I(v)
    = \frac{1}{|P|}\sum_{j\in P}\frac{1}{\rank_j(v)}.
    \label{eq:dmrr}
\end{equation}

Equation~\eqref{eq:dmrr} depends only on within-position vocabulary rankings and thus does not require score magnitudes to be comparable across visual positions.
Reciprocal rank gives greater weight to positions where $v$ ranks near the top while limiting the contribution of low-ranked positions.
The quantity $\dmrr_I(v)$ depends on the image and readout layer but is independent of the autoregressive step.
We therefore compute it once during prefill and cache it for the entire response.

\begin{table*}[t]
    \centering
    \footnotesize
    \renewcommand{\arraystretch}{0.90}
    \setlength{\tabcolsep}{2.6pt}
    \begin{tabular}{lcccccccccccc}
        \toprule
        & \multicolumn{3}{c}{CHAIR}
        & \multicolumn{7}{c}{AMBER}
        & \multicolumn{2}{c}{Cost} \\
        \cmidrule(lr){2-4}\cmidrule(lr){5-11}\cmidrule(lr){12-13}
        Method
        & CHAIR$_S\downarrow$ & CHAIR$_I\downarrow$ & F1$\uparrow$
        & CHAIR$\downarrow$ & Cover$\uparrow$ & Hal.$\downarrow$ & Cog.$\downarrow$
        & Acc.$\uparrow$ & F1$\uparrow$
        & Score$\uparrow$
        & Lat.$\downarrow$ & Mem.$\downarrow$ \\
        \midrule
        \llavaonefive
        & 52.60 & 15.61 & 80.66 & \underline{3.69} & 50.47 & 19.02 & 3.82 & 71.96 & 74.81 & 85.56 & 1.00$\times$ & 1.00$\times$ \\
        \quad VCD
        & \cellcolor{red!15}59.00 & \cellcolor{red!15}17.97 & 78.50 & \cellcolor{red!35}6.25 & \underline{51.81} & \cellcolor{red!35}32.07 & \cellcolor{red!15}4.49 & \cellcolor{red!15}66.81 & \cellcolor{red!15}70.37 & 82.06 & \cellcolor{red!35}2.01$\times$ & \cellcolor{red!15}1.06$\times$ \\
        \quad OPERA
        & 51.20 & \cellcolor{green!15}14.19 & \textbf{80.96} & 3.70 & 48.79 & \cellcolor{green!15}\underline{17.03} & \cellcolor{green!35}\underline{2.70} & \textbf{74.94} & \textbf{78.09} & \textbf{87.20} & \cellcolor{red!35}7.26$\times$ & \cellcolor{red!35}1.51$\times$ \\
        \quad DoLa
        & 53.00 & 15.87 & 79.28 & \cellcolor{red!35}4.52 & 51.03 & \cellcolor{red!15}21.12 & 3.84 & 72.17 & 75.09 & 85.28 & \cellcolor{red!15}1.10$\times$ & 1.00$\times$ \\
        \quad PAI
        & \cellcolor{green!15}\textbf{42.40} & \cellcolor{green!15}\underline{13.51} & \cellcolor{red!15}75.55 & \cellcolor{red!35}4.68 & \cellcolor{red!15}43.11 & 19.32 & \cellcolor{green!35}\textbf{1.97} & \cellcolor{red!35}38.29 & \cellcolor{red!35}18.93 & \cellcolor{red!35}57.12 & \cellcolor{red!35}2.14$\times$ & 1.02$\times$ \\
        \quad ReVisiT
        & 51.40 & 15.44 & 80.57 & \cellcolor{red!15}4.00 & \cellcolor{green!15}\textbf{55.15} & \cellcolor{red!35}24.90 & \cellcolor{green!35}2.72 & \underline{73.09} & \underline{77.80} & \underline{86.90} & 1.02$\times$ & 1.03$\times$ \\
        \quad \textbf{\methodname}
        & \cellcolor{green!15}\underline{44.80} & \cellcolor{green!15}\textbf{12.67} & \underline{80.85} & \cellcolor{green!15}\textbf{2.98} & 50.21 & \cellcolor{green!15}\textbf{16.43} & \cellcolor{green!15}3.13 & 71.96 & 74.81 & 85.91 & 1.02$\times$ & 1.00$\times$ \\
        \midrule
        \qwenvlseven
        & 31.60 & 9.58 & \underline{70.80} & 4.52 & 55.80 & 20.32 & \underline{1.02} & 77.85 & 87.89 & 91.68 & 1.00$\times$ & 1.00$\times$ \\
        \quad VCD
        & \cellcolor{red!15}33.80 & 9.65 & 70.27 & \cellcolor{red!35}5.43 & \cellcolor{green!15}\underline{63.30} & \cellcolor{red!35}26.99 & \cellcolor{red!35}1.77 & \cellcolor{green!15}81.77 & 86.66 & 90.61 & \cellcolor{red!35}2.03$\times$ & 1.01$\times$ \\
        \quad OPERA
        & \cellcolor{green!35}\textbf{20.20} & \cellcolor{green!35}\underline{7.01} & \cellcolor{red!15}63.79 & \cellcolor{green!35}\textbf{3.51} & \cellcolor{green!15}\textbf{63.33} & 20.92 & \cellcolor{red!35}1.42 & \cellcolor{green!15}\textbf{83.70} & \textbf{88.17} & \textbf{92.33} & \cellcolor{red!35}8.03$\times$ & 1.04$\times$ \\
        \quad DoLa
        & \cellcolor{green!35}\underline{23.60} & \cellcolor{red!35}16.76 & \cellcolor{red!15}60.32 & 4.47 & \cellcolor{red!15}51.71 & \cellcolor{green!15}18.92 & \cellcolor{red!35}1.56 & \cellcolor{green!15}82.55 & 87.15 & 91.33 & \cellcolor{red!35}1.25$\times$ & 1.00$\times$ \\
        \quad PAI
        & \cellcolor{green!15}26.20 & 9.61 & \cellcolor{red!15}66.98 & \cellcolor{green!15}4.25 & \cellcolor{red!15}49.47 & \cellcolor{green!15}\textbf{18.03} & \cellcolor{red!15}1.10 & 81.34 & 86.26 & 91.00 & \cellcolor{red!35}2.04$\times$ & 1.00$\times$ \\
        \quad ReVisiT
        & \cellcolor{green!35}24.00 & \cellcolor{green!35}\textbf{6.75} & \cellcolor{red!15}66.99 & \cellcolor{green!15}4.27 & \cellcolor{green!15}62.30 & \cellcolor{red!35}24.90 & \cellcolor{red!35}1.42 & \cellcolor{green!15}\underline{83.28} & 87.70 & 91.71 & 1.05$\times$ & \cellcolor{red!35}1.25$\times$ \\
        \quad \textbf{\methodname}
        & \cellcolor{green!15}25.40 & \cellcolor{green!35}7.54 & \textbf{71.83} & \cellcolor{green!15}\underline{3.80} & 54.65 & \cellcolor{green!15}\underline{18.82} & \cellcolor{green!15}\textbf{0.93} & 78.19 & \underline{87.97} & \underline{92.08} & 1.01$\times$ & 1.00$\times$ \\
        \midrule
        \instructblipseven
        & 49.60 & \underline{14.03} & 79.01 & 4.64 & \underline{53.74} & 24.40 & 3.80 & \underline{75.91} & \underline{81.40} & \underline{88.38} & 1.00$\times$ & 1.00$\times$ \\
        \quad VCD
        & \cellcolor{red!15}56.20 & \cellcolor{red!35}17.67 & 76.48 & \cellcolor{red!35}7.25 & 53.24 & \cellcolor{red!35}36.16 & \cellcolor{red!35}4.59 & \cellcolor{red!15}70.23 & \cellcolor{red!15}75.94 & 84.34 & \cellcolor{red!35}1.98$\times$ & 1.02$\times$ \\
        \quad OPERA
        & 50.80 & \cellcolor{red!15}14.77 & \underline{79.02} & 4.81 & 52.08 & 23.80 & \cellcolor{green!15}3.53 & 75.90 & 81.32 & 88.25 & \cellcolor{red!35}11.52$\times$ & \cellcolor{red!15}1.18$\times$ \\
        \quad DoLa
        & \cellcolor{red!35}67.00 & \cellcolor{red!35}20.26 & 76.84 & \cellcolor{red!35}7.14 & \textbf{55.26} & \cellcolor{red!35}35.86 & \cellcolor{red!35}7.09 & 75.76 & 81.06 & 86.96 & \cellcolor{red!35}1.21$\times$ & 1.00$\times$ \\
        \quad PAI
        & \cellcolor{red!35}61.00 & \cellcolor{red!15}16.70 & 77.79 & 4.63 & 52.03 & \cellcolor{green!15}\underline{22.81} & \cellcolor{red!15}4.04 & 75.30 & 80.03 & 87.70 & \cellcolor{red!35}2.19$\times$ & 1.01$\times$ \\
        \quad ReVisiT
        & \cellcolor{green!15}\textbf{43.00} & \cellcolor{red!15}15.02 & \cellcolor{red!15}74.97 & \cellcolor{green!15}\underline{4.22} & \cellcolor{red!15}50.99 & \cellcolor{green!15}22.91 & \cellcolor{green!35}\textbf{2.13} & \cellcolor{red!35}54.36 & \cellcolor{red!35}50.95 & \cellcolor{red!15}73.37 & 1.01$\times$ & 1.02$\times$ \\
        \quad \textbf{\methodname}
        & \cellcolor{green!15}\underline{46.00} & \cellcolor{green!15}\textbf{12.50} & \textbf{79.80} & \cellcolor{green!15}\textbf{4.05} & 53.28 & \cellcolor{green!15}\textbf{21.51} & \cellcolor{green!15}\underline{3.11} & \textbf{75.96} & \textbf{81.44} & \textbf{88.69} & 1.01$\times$ & 1.00$\times$ \\
        \bottomrule
    \end{tabular}
    \caption{\textbf{Main results on CHAIR and AMBER with decoding cost.} The
    AMBER columns include generative and discriminative metrics, followed by
    the benchmark's combined score. Lat.\ and Mem.\ are per-token latency
    and peak allocated memory relative to each backbone's greedy decoding
    under the cached-evidence protocol. Green and red shade improvements
    and degradations of more than 5\% against each
    backbone's greedy baseline, darker shades mark more than 20\%, and
    unshaded cells differ by less than 5\%. Bold and underlined values
    indicate the best and second-best result, respectively, within each
    backbone block. Arrows indicate the preferred direction.}
    \label{tab:chair-amber}
\end{table*}

\subsection{Register: Uncertainty-Aware Token Calibration}
\label{sec:calib}

The Register module asks whether a token-specific reference is reliable enough to drive an intervention.
We run the base model without logit intervention on $N$ unlabeled calibration images that are disjoint from the evaluation data.
For calibration image $I_i$ with $T_i$ decoding steps, let $\Cand_{i,t}$ be the candidate set obtained by top-$p$ selection under the base-model distribution, subject to minimum and maximum sizes.
We define the candidate-conditioned calibration multiset as
\begin{equation}
    D_v
    =
    \biguplus_{\substack{1\leq i\leq N,\;1\leq t\leq T_i\\
                         v\in\Cand_{i,t}}}
    \left\{\dmrr_{I_i}(v)\right\},
    \label{eq:calibration-multiset}
\end{equation}
where $\biguplus$ denotes multiset union and therefore preserves repeated candidate occurrences.
Thus, $D_v$ records the visual evidence for $v$ only at steps where the model considers it a plausible next-token candidate.

We summarize the observations using a token-specific reference and a pooled normalization scale:
\begin{equation}
    \begin{aligned}
        b(v) &= \operatorname{median}(D_v), \\
        b_0  &= \operatorname{median}\!\left(\biguplus_v D_v\right),
    \end{aligned}
    \label{eq:baseline}
\end{equation}
The token reference $b(v)$ estimates the typical evidence level of token $v$ as a decoder candidate under the calibration distribution (Section~\ref{sec:offset}), whereas $b_0$ provides a shared normalization scale across tokens.

Tokens occur at different frequencies during calibration, so the reliability of $b(v)$ also varies.
To account for this uncertainty, we construct a nominal 95\% order-statistic range over the candidate-occurrence multiset $D_v$, denoted by $[b_{\mathrm{lo}}(v),b_{\mathrm{hi}}(v)]$. We use it as a conservative stability diagnostic rather than a coverage-guaranteed confidence interval (Appendix~\ref{sec:registration}).
We then define the normalized edit induced by a reference $b$ and observation $x$ as
\begin{equation}
    e(b,x)=\clip\!\left(\frac{b-x}{b_0},0,1\right).
    \label{eq:normalized-edit}
\end{equation}
We quantify estimation uncertainty by its effect on the bounded edit:
\begin{equation}
    \Delta e(v)
    =\frac{1}{|D_v|}\sum_{x\in D_v}
    \left|e\!\left(b_{\mathrm{hi}}(v),x\right)
          -e\!\left(b_{\mathrm{lo}}(v),x\right)\right|.
    \label{eq:deltae}
\end{equation}
We register token $v$ only if its range exists and $\Delta e(v)<0.5$.
Otherwise, the method abstains from editing it.
The frozen table $\mathcal T$ stores the global scale $b_0$ and maps each registered vocabulary token $v\in\mathcal R$ to its token-specific reference $b(v)$.
Appendix~\ref{sec:registration} provides the range construction and Appendix~\ref{sec:implementation-details} the candidate-set configuration.

\subsection{Intervene: Bounded Evidence-Deficit Suppression}
\label{sec:suppress}

The Intervene module determines how strongly to suppress a candidate that lacks visual support.
For a registered token, we reuse Eq.~\eqref{eq:normalized-edit} to compute an image-specific suppression strength
\begin{equation}
    \begin{aligned}
        s_I(v) &= e\!\left(b(v),\dmrr_I(v)\right).
    \end{aligned}
    \label{eq:suppression-strength}
\end{equation}
At decoding step $t$, let $\mathbf z_t\in\mathbb R^{|\vocab|}$ denote the decoder logits before intervention, $z_t(v)$ their component for token $v\in\vocab$, and $\Cand_t$ the candidate set constructed exactly as during calibration.
Given maximum penalty $\beta\geq0$, the edited decoder logit is
\begin{equation}
    z'_t(v)=
    \begin{cases}
        z_t(v)-\beta s_I(v), & v\in\Cand_t\cap\mathcal R,\\[3pt]
        z_t(v),              & \text{otherwise}.
    \end{cases}
    \label{eq:shift}
\end{equation}

\begin{table*}[t]
    \centering
    \footnotesize
    \renewcommand{\arraystretch}{0.93}
    \setlength{\tabcolsep}{5pt}
    \begin{tabular}{llcccccc>{\columncolor{gray!15}}c}
        \toprule
        Backbone & Metric
        & Base & VCD & OPERA & DoLa & PAI & ReVisiT & \methodname \\
        \midrule
        \llavaonefive & MMHal Score$\uparrow$
        & \underline{2.76} & 2.16 & 2.57 & 2.65 & 2.28 & 2.33 & \textbf{2.78} \\
        & MMHal Hall. Rate (\%)$\downarrow$
        & 51.04 & 65.62 & 53.12 & 53.12 & \textbf{43.75} & 58.33 & \underline{48.96} \\
        \cmidrule(l){2-9}
        & MM-Vet Acc.$\uparrow$
        & 35.41 & 32.84 & \underline{36.74} & 36.06 & 18.26 & 33.58 & \textbf{37.20} \\
        \midrule
        \qwenvlseven & MMHal Score$\uparrow$
        & 4.28 & 4.11 & \textbf{4.50} & 3.69 & 3.27 & 3.78 & \underline{4.42} \\
        & MMHal Hall. Rate (\%)$\downarrow$
        & \textbf{15.62} & \underline{17.71} & \underline{17.71} & 20.83 & 26.04 & 19.79 & \textbf{15.62} \\
        \cmidrule(l){2-9}
        & MM-Vet Acc.$\uparrow$
        & 56.65 & \underline{60.87} & \textbf{70.96} & 46.97 & 44.86 & 56.38 & 59.54 \\
        \midrule
        \instructblipseven & MMHal Score$\uparrow$
        & 2.34 & 2.19 & 2.34 & 2.33 & \underline{2.36} & 2.00 & \textbf{2.54} \\
        & MMHal Hall. Rate (\%)$\downarrow$
        & 55.21 & 60.42 & 55.21 & \underline{54.17} & \underline{54.17} & 60.42 & \textbf{48.96} \\
        \cmidrule(l){2-9}
        & MM-Vet Acc.$\uparrow$
        & 29.17 & 30.78 & \textbf{35.23} & \underline{31.61} & 31.19 & 20.09 & 29.63 \\
        \midrule
        \llavanextseven & MMHal Score$\uparrow$
        & 3.84 & 3.46 & 3.71 & \underline{3.96} & 3.55 & 3.47 & \textbf{4.02} \\
        & MMHal Hall. Rate (\%)$\downarrow$
        & \underline{31.25} & 41.67 & 37.50 & 32.29 & 38.54 & 37.50 & \textbf{29.17} \\
        \cmidrule(l){2-9}
        & MM-Vet Acc.$\uparrow$
        & 48.21 & 46.19 & \underline{52.02} & \textbf{52.43} & 49.50 & 49.31 & 50.05 \\
        \bottomrule
    \end{tabular}
    \caption{\textbf{MMHal-Bench and MM-Vet results.} Hall. Rate is the
    percentage of examples with score below 3, and MM-Vet accuracy is on a
    0--100 scale.}
    \label{tab:mmhal-bench}
\end{table*}

The lower clipping bound makes the update suppression-only.
Evidence at or above the token reference yields no edit, whereas lower evidence can reduce a candidate's logit by at most $\beta$.
Candidates without registered references and tokens outside $\Cand_t$ remain unchanged.
At each decoding step, we recompute the candidate set from the pre-edit distribution.
In contrast, we compute $s_I$ once per image and cache it throughout decoding.
Overall, online inference combines the frozen calibration table $\mathcal T$ with a single DMRR computation at prefill, after which each decoding step requires only candidate selection and bounded logit updates.
Appendix~\ref{sec:mechanism-case} traces this arithmetic at a recorded decoding step, where full, proportional, and zero suppression occur within a single candidate set.

\section{Experiments}
\label{sec:experiments}

\subsection{Experimental Setup}
\label{sec:setup}

\paragraph{Models.}
We evaluate four 7B LVLMs with different visual interfaces and language backbones, namely \llavaonefive{} \citep{liu2024improved}, \qwenvlseven{} \citep{bai2025qwen25vl}, \instructblipseven{} \citep{dai2023instructblip}, and \llavanextseven{} \citep{liu2024llavanext}.
Appendix~\ref{sec:implementation-details} lists the exact checkpoint identifiers.

% Row names follow the Register/Intervene component terms of Section 4;
% arm definitions are in table/A16_Definitions.tex.
\begin{table}[t]
    \centering
    \setlength{\tabcolsep}{2pt}
    \resizebox{0.96\columnwidth}{!}{%
    \begin{tabular}{l*{6}{>{\centering\arraybackslash}m{1.05cm}}}
        \toprule
        \multirow{2}{*}{Configuration} & \multicolumn{3}{c}{CHAIR} & \multicolumn{3}{c}{AMBER} \\
        \cmidrule(lr){2-4}
        \cmidrule(lr){5-7}
        & $C_S\,\downarrow$
        & $C_I\,\downarrow$
        & F1 $\uparrow$
        & \multicolumn{1}{c}{CHAIR $\downarrow$}
        & \multicolumn{1}{c}{Cover $\uparrow$}
        & \multicolumn{1}{c}{Score $\uparrow$} \\
        \midrule
        \rowcolor{gray!15}[3pt][3pt]
        Full method             & 44.8 & 12.67 & 80.85 & 2.98 & 50.21 & 85.91 \\
        Global reference        & 58.4 & 17.45 & 80.31 & 4.10 & 52.78 & 85.36 \\
        Shuffled references     & 56.0 & 14.74 & 81.54 & 4.11 & 51.63 & 85.36 \\
        Mismatched evidence     & 50.2 & 14.45 & 79.73 & 3.70 & 49.35 & 85.56 \\
        No abstention           & 44.4 & 12.64 & 80.90 & 2.97 & 50.37 & 86.62 \\
        Bidirectional update    & 50.4 & 16.07 & 79.74 & 3.72 & 51.22 & 85.54 \\
        Unbounded update        & 39.2 & 29.71 & 69.09 & 3.44 & 48.26 & 85.69 \\
        \bottomrule
    \end{tabular}
    }
    \caption{\textbf{Component ablations on \llavaonefive{}.} $C_S$ and
    $C_I$ denote CHAIR$_S$ and CHAIR$_I$, and arrows indicate the preferred
    direction. Arm definitions are in Table~\ref{tab:ablation-definitions}.}
    \label{tab:signal-ablation}
\end{table}

\paragraph{Benchmarks.}
CHAIR \citep{rohrbach2018object} evaluates object hallucination in MS COCO captions.
AMBER \citep{wang2023amber} provides generative and discriminative evaluations.
MMHal-Bench \citep{sun2023factually} evaluates open-ended multimodal responses.
We use MM-Vet \citep{yu2024mmvet} to measure general multimodal capability.
Because a model can reduce hallucination simply by saying less, we read hallucination metrics together with recall-oriented measures and general capability.
Appendix~\ref{sec:evaluation-prompts} gives the splits, prompts, metrics, and evaluator contracts.

\paragraph{Implementation.}
We calibrate on 500 MS COCO training images and define the candidate set as a top-$p$ probability prefix with $p=0.9$, restricted to between 2 and 50 tokens.
We select the readout layer and $\beta$ once per backbone on 500 images
sampled from MS COCO train2014 with seed 42. We then freeze the selected
setting across benchmarks. CHAIR results come from a separate 500-image
subset of MS COCO val2014, so no evaluation image enters operating-point
selection.
We compare with VCD \citep{leng2024vcd}, OPERA \citep{huang2024opera}, DoLa \citep{chuang2024dola}, PAI \citep{liu2024pai}, and ReVisiT \citep{cho-kim-2026-revisit}.
Appendix~\ref{sec:implementation-details} provides further details.

\begin{figure*}[t]
    \centering
    \includegraphics[width=0.98\textwidth]{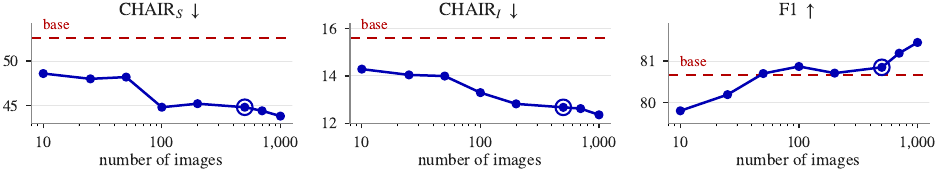}
    \caption{\textbf{Calibration-size sweep on \llavaonefive{}.} Dashed
    lines mark base decoding, and the open circle marks the deployed
    500-image table.}
    \label{fig:calibration-size}
\end{figure*}

\subsection{Main Results}
\label{sec:results}

\paragraph{Visual Grounding and Hallucination Mitigation.}
Table~\ref{tab:chair-amber} and Appendix~\ref{sec:additional-results} summarize fine-grained visual grounding and hallucination mitigation on CHAIR and AMBER.
Across the four models, \methodname{} reduces CHAIR$_I$ by $10.3\%$--$21.3\%$ while largely preserving or improving F1, so it does not simply suppress object mentions.
The improvement generalizes to 11 models spanning six architecture families and model sizes from 7B to 32B, and \methodname{} reduces both CHAIR$_S$ and CHAIR$_I$ for every model (Appendix~\ref{sec:model-scaling}).

On AMBER generation, \methodname{} lowers CHAIR and hallucination rate and improves the AMBER score for all four backbones.
OPERA reduces CHAIR$_S$ more strongly on Qwen, and PAI reaches a lower CHAIR$_S$ on LLaVA-1.5, but both give up more F1 or discriminative performance.
\methodname{} balances hallucination reduction and retained content more consistently.
Under the shading rule of Table~\ref{tab:chair-amber}, \methodname{} is the only method that never degrades a baseline metric by more than $5\%$ on any backbone.
Its largest regression is a $2.1\%$ relative decrease in AMBER coverage.

\paragraph{Open-ended Reliability and General Multimodal Utility.}

Table~\ref{tab:mmhal-bench} shows that \methodname{} improves MMHal-Bench quality and MM-Vet accuracy across all four backbones, while reducing or maintaining the MMHal-Bench hallucination rate.

\subsection{Component and Counterfactual Analysis}
\label{sec:ablation-study}

\paragraph{Impact of the Ordinal Readout.}
Under a matched mean edit budget, replacing DMRR with probability pooling in Table~\ref{tab:detection-not-mitigation} increases CHAIR$_S$ from $44.8$ to $50.0$.
This comparison supports the ordinal readout as part of the complete intervention.

\paragraph{Impact of Token-Specific Calibration.}
Table~\ref{tab:signal-ablation} shows that global or shuffled references degrade performance.
Because shuffling preserves the marginal reference distribution, the degradation isolates the correspondence between tokens and their references.

\paragraph{Impact of Calibration Size.}
Figure~\ref{fig:calibration-size} refits the calibration table on nested COCO subsets from 10 to 1,000 images at the frozen operating point.
Ten images already recover half of the CHAIR$_S$ reduction, and 100 images match the default CHAIR$_S$ while keeping F1 above base decoding.
We fixed the 500-image calibration table before these runs, so the reported results do not sit at this sweep's optimum (Appendix~\ref{sec:design-space}).

\paragraph{Impact of Image-Specific Evidence.}
The mismatched-evidence arm in Table~\ref{tab:signal-ablation} rotates the cached evidence vector across images while holding every other component fixed.
Appendix~\ref{sec:additional-results} extends this counterfactual with further substitution variants (Table~\ref{tab:image-specificity}).
Every substitution recovers less than half of the full method's CHAIR$_S$ reduction and pushes F1 below base decoding, so the intervention draws on evidence about the current image, not only on which tokens are registered.

\paragraph{Impact of Conservative Suppression.}
Table~\ref{tab:signal-ablation} shows that bidirectional updates degrade CHAIR$_S$, CHAIR$_I$, and F1.
Unbounded updates lower CHAIR$_S$ but induce repetition, which sharply increases CHAIR$_I$ and reduces F1.
In contrast, removing abstention has little effect on the aggregate metrics and affects only about $3.5\%$ of candidate positions.
We therefore treat registration as a conservative reference-stability rule rather than a measured source of aggregate gains.

\subsection{Inference Efficiency}
\label{sec:efficiency}

The Lat.\ and Mem.\ columns of Table~\ref{tab:chair-amber} report per-token decoding latency and peak allocated memory relative to greedy decoding, measured separately on three backbones over 100 prompts each and excluding offline calibration.
Across the three backbones, \methodname{} adds $1.33\%$ latency after the evidence has been cached. Including online DMRR computation during prefill yields $2.40\%$ end-to-end latency overhead.
Peak allocated memory increases by $0.31\%$ under the cached-evidence protocol.
Appendix~\ref{sec:computational-cost} details both latency paths and the evidence-cache footprint.
\section{Conclusion}
\label{sec:conclusion}

We introduced \methodname{}, a training-free decoder that aggregates vocabulary ranks across visual positions and compares each candidate with a reliably estimated token-specific reference.
A bounded, one-sided penalty converts evidence deficits into logit edits, and prefill-time computation lets generation reuse the evidence.
Across four 7B backbones, \methodname{} reduces hallucination while preserving or improving descriptive and general multimodal utility, and the reductions extend to 11 models across six architecture families from 7B to 32B.
These results support token-calibrated ordinal readouts as a practical interface between internal visual states and generation.

\section*{Limitations}

\methodname{} requires access to visual-token hidden states, the output normalization, and the vocabulary head, so it cannot operate through closed APIs.
Our current implementation uses a separate calibration table, readout layer, and intervention strength for each backbone.
We evaluate 11 models from six architecture families, but we calibrate and tune each model separately.
Our experiments therefore do not establish whether these components transfer across backbones without additional calibration or development data.
Future work could derive references that transfer across models and choose operating points automatically.

The method derives its corrective signal entirely from the model's visual representations and does not consult external sources.
It can lower the logits of candidates that the image does not support, but it cannot add facts that neither the image nor the backbone contains.
It therefore offers no direct way to correct errors that require current or specialized knowledge.
Our open-ended and general multimodal benchmarks do not test this type of factual error.
Future work could combine ordinal visual evidence with retrieval or external verification.

We conduct all calibration, development, and evaluation in English.
We construct token-specific references over each model's vocabulary, so differences in multilingual tokenization and alignment may change candidate frequencies and visual-evidence baselines.
Our results do not show whether the method behaves similarly in other languages, especially when languages differ in morphology or tokenization.
We could repeat calibration in another language, but we do not know whether a reference table can transfer across languages or requires separate calibration.
Future work should test the method in multilingual settings.

\section*{Ethical Considerations}

Reducing visually unsupported statements can improve the reliability of LVLM
outputs, but the method does not make a model safe for high-stakes use. A
bounded intervention can still leave hallucinations unchanged or replace them
with different errors. Users should independently verify claims in medical,
legal, scientific, and public-safety settings.

The method requires access to internal model states and an offline calibration
corpus. Deployments should respect the licenses, privacy constraints, and
provenance requirements of both model checkpoints and calibration images.
Calibration data can also encode geographic, cultural, and frequency biases.
Token-specific references may reproduce these biases by treating common visual
concepts as better calibrated than rare ones. We report abstention for
unstable references and avoid presenting lower benchmark scores as evidence
of universal factual reliability.

\bibliography{custom}

@inproceedings{rohrbach2018object,
  title     = {Object Hallucination in Image Captioning},
  author    = {Rohrbach, Anna and Hendricks, Lisa Anne and Burns, Kaylee and
               Darrell, Trevor and Saenko, Kate},
  booktitle = {Proceedings of the 2018 Conference on Empirical Methods in
               Natural Language Processing},
  pages     = {4035--4045},
  month     = {October--November},
  address   = {Brussels, Belgium},
  publisher = {Association for Computational Linguistics},
  doi       = {10.18653/v1/D18-1437},
  url       = {https://aclanthology.org/D18-1437/},
  year      = {2018}
}

@inproceedings{lin2014coco,
  title     = {{Microsoft} {COCO}: Common Objects in Context},
  author    = {Lin, Tsung-Yi and Maire, Michael and Belongie, Serge and Hays,
               James and Perona, Pietro and Ramanan, Deva and Doll{\'a}r, Piotr
               and Zitnick, C. Lawrence},
  booktitle = {Computer Vision -- ECCV 2014},
  editor    = {Fleet, David and Pajdla, Tomas and Schiele, Bernt and
               Tuytelaars, Tinne},
  series    = {Lecture Notes in Computer Science},
  volume    = {8693},
  pages     = {740--755},
  address   = {Zurich, Switzerland},
  publisher = {Springer},
  doi       = {10.1007/978-3-319-10602-1_48},
  url       = {https://doi.org/10.1007/978-3-319-10602-1_48},
  year      = {2014}
}

@inproceedings{hudson2019gqa,
  title     = {{GQA}: A New Dataset for Real-World Visual Reasoning and
               Compositional Question Answering},
  author    = {Hudson, Drew A. and Manning, Christopher D.},
  booktitle = {Proceedings of the IEEE/CVF Conference on Computer Vision and
               Pattern Recognition (CVPR)},
  pages     = {6700--6709},
  month     = {June},
  address   = {Long Beach, CA, USA},
  publisher = {IEEE},
  doi       = {10.1109/CVPR.2019.00686},
  url       = {https://openaccess.thecvf.com/content_CVPR_2019/html/Hudson_GQA_A_New_Dataset_for_Real-World_Visual_Reasoning_and_Compositional_CVPR_2019_paper.html},
  year      = {2019}
}

@misc{wang2023amber,
  title     = {{AMBER}: An {LLM}-free Multi-dimensional Benchmark for {MLLMs}
               Hallucination Evaluation},
  author    = {Wang, Junyang and Wang, Yuhang and Xu, Guohai and Zhang, Jing
               and Gu, Yukai and Jia, Haitao and Wang, Jiaqi and Xu, Haiyang
               and Yan, Ming and Zhang, Ji and Sang, Jitao},
  eprint    = {2311.07397},
  archiveprefix = {arXiv},
  doi       = {10.48550/arXiv.2311.07397},
  url       = {https://arxiv.org/abs/2311.07397},
  year      = {2023}
}

@inproceedings{sun2023factually,
  title     = {Aligning Large Multimodal Models with Factually Augmented {RLHF}},
  author    = {Sun, Zhiqing and Shen, Sheng and Cao, Shengcao and Liu, Haotian
               and Li, Chunyuan and Shen, Yikang and Gan, Chuang and Gui,
               Liangyan and Wang, Yu-Xiong and Yang, Yiming and Keutzer, Kurt
               and Darrell, Trevor},
  booktitle = {Findings of the Association for Computational Linguistics: ACL
               2024},
  pages     = {13088--13110},
  month     = {August},
  address   = {Bangkok, Thailand},
  publisher = {Association for Computational Linguistics},
  doi       = {10.18653/v1/2024.findings-acl.775},
  url       = {https://aclanthology.org/2024.findings-acl.775/},
  year      = {2024}
}

@inproceedings{yu2024mmvet,
  title     = {{MM-Vet}: Evaluating Large Multimodal Models for Integrated
               Capabilities},
  author    = {Yu, Weihao and Yang, Zhengyuan and Li, Linjie and Wang, Jianfeng
               and Lin, Kevin and Liu, Zicheng and Wang, Xinchao and Wang,
               Lijuan},
  booktitle = {Proceedings of the 41st International Conference on Machine
               Learning},
  series    = {Proceedings of Machine Learning Research},
  volume    = {235},
  pages     = {57730--57754},
  address   = {Vienna, Austria},
  publisher = {PMLR},
  url       = {https://proceedings.mlr.press/v235/yu24o.html},
  year      = {2024}
}

@inproceedings{liu2023llava,
  title     = {Visual Instruction Tuning},
  author    = {Liu, Haotian and Li, Chunyuan and Wu, Qingyang and Lee, Yong Jae},
  booktitle = {Advances in Neural Information Processing Systems},
  editor    = {Oh, Alice and Naumann, Tristan and Globerson, Amir and Saenko,
               Kate and Hardt, Moritz and Levine, Sergey},
  volume    = {36},
  pages     = {34892--34916},
  publisher = {Curran Associates, Inc.},
  doi       = {10.52202/075280-1516},
  url       = {https://proceedings.neurips.cc/paper_files/paper/2023/file/6dcf277ea32ce3288914faf369fe6de0-Paper-Conference.pdf},
  year      = {2023}
}

@inproceedings{liu2024improved,
  title     = {Improved Baselines with Visual Instruction Tuning},
  author    = {Liu, Haotian and Li, Chunyuan and Li, Yuheng and Lee, Yong Jae},
  booktitle = {Proceedings of the IEEE/CVF Conference on Computer Vision and
               Pattern Recognition (CVPR)},
  pages     = {26296--26306},
  month     = {June},
  address   = {Seattle, WA, USA},
  publisher = {IEEE},
  doi       = {10.1109/CVPR52733.2024.02484},
  url       = {https://openaccess.thecvf.com/content/CVPR2024/html/Liu_Improved_Baselines_with_Visual_Instruction_Tuning_CVPR_2024_paper.html},
  year      = {2024}
}

@inproceedings{dai2023instructblip,
  title     = {{InstructBLIP}: Towards General-purpose Vision-Language Models
               with Instruction Tuning},
  author    = {Dai, Wenliang and Li, Junnan and Li, Dongxu and Tiong, Anthony
               and Zhao, Junqi and Wang, Weisheng and Li, Boyang and Fung,
               Pascale N. and Hoi, Steven C.},
  booktitle = {Advances in Neural Information Processing Systems},
  editor    = {Oh, Alice and Naumann, Tristan and Globerson, Amir and Saenko,
               Kate and Hardt, Moritz and Levine, Sergey},
  volume    = {36},
  pages     = {49250--49267},
  publisher = {Curran Associates, Inc.},
  doi       = {10.52202/075280-2142},
  url       = {https://proceedings.neurips.cc/paper_files/paper/2023/file/9a6a435e75419a836fe47ab6793623e6-Paper-Conference.pdf},
  year      = {2023}
}

@misc{bai2025qwen25vl,
  title     = {{Qwen2.5-VL} Technical Report},
  author    = {Bai, Shuai and Chen, Keqin and Liu, Xuejing and Wang, Jialin and
               Ge, Wenbin and Song, Sibo and Dang, Kai and Wang, Peng and Wang,
               Shijie and Tang, Jun and Zhong, Humen and Zhu, Yuanzhi and Yang,
               Mingkun and Li, Zhaohai and Wan, Jianqiang and Wang, Pengfei and
               Ding, Wei and Fu, Zheren and Xu, Yiheng and Ye, Jiabo and Zhang,
               Xi and Xie, Tianbao and Cheng, Zesen and Zhang, Hang and Yang,
               Zhibo and Xu, Haiyang and Lin, Junyang},
  eprint    = {2502.13923},
  archiveprefix = {arXiv},
  doi       = {10.48550/arXiv.2502.13923},
  url       = {https://arxiv.org/abs/2502.13923},
  year      = {2025}
}

@inproceedings{leng2024vcd,
  title     = {Mitigating Object Hallucinations in Large Vision-Language Models
               through Visual Contrastive Decoding},
  author    = {Leng, Sicong and Zhang, Hang and Chen, Guanzheng and Li, Xin and
               Lu, Shijian and Miao, Chunyan and Bing, Lidong},
  booktitle = {Proceedings of the IEEE/CVF Conference on Computer Vision and
               Pattern Recognition (CVPR)},
  pages     = {13872--13882},
  month     = {June},
  address   = {Seattle, WA, USA},
  publisher = {IEEE},
  doi       = {10.1109/CVPR52733.2024.01316},
  url       = {https://openaccess.thecvf.com/content/CVPR2024/html/Leng_Mitigating_Object_Hallucinations_in_Large_Vision-Language_Models_through_Visual_Contrastive_CVPR_2024_paper.html},
  year      = {2024}
}

@inproceedings{huang2024opera,
  title     = {{OPERA}: Alleviating Hallucination in Multi-Modal Large Language
               Models via Over-Trust Penalty and Retrospection-Allocation},
  author    = {Huang, Qidong and Dong, Xiaoyi and Zhang, Pan and Wang, Bin and
               He, Conghui and Wang, Jiaqi and Lin, Dahua and Zhang, Weiming
               and Yu, Nenghai},
  booktitle = {Proceedings of the IEEE/CVF Conference on Computer Vision and
               Pattern Recognition (CVPR)},
  pages     = {13418--13427},
  month     = {June},
  address   = {Seattle, WA, USA},
  publisher = {IEEE},
  doi       = {10.1109/CVPR52733.2024.01274},
  url       = {https://openaccess.thecvf.com/content/CVPR2024/html/Huang_OPERA_Alleviating_Hallucination_in_Multi-Modal_Large_Language_Models_via_Over-Trust_CVPR_2024_paper.html},
  year      = {2024}
}

@inproceedings{chuang2024dola,
  title     = {{DoLa}: Decoding by Contrasting Layers Improves Factuality in
               Large Language Models},
  author    = {Chuang, Yung-Sung and Xie, Yujia and Luo, Hongyin and Kim, Yoon
               and Glass, James R. and He, Pengcheng},
  booktitle = {The Twelfth International Conference on Learning Representations},
  url       = {https://openreview.net/forum?id=Th6NyL07na},
  year      = {2024}
}

@inproceedings{liu2024pai,
  title     = {Paying More Attention to Image: A Training-Free Method for
               Alleviating Hallucination in {LVLMs}},
  author    = {Liu, Shi and Zheng, Kecheng and Chen, Wei},
  booktitle = {Computer Vision -- ECCV 2024},
  editor    = {Leonardis, Ale{\v{s}} and Ricci, Elisa and Roth, Stefan and
               Russakovsky, Olga and Sattler, Torsten and Varol, G{\"u}l},
  series    = {Lecture Notes in Computer Science},
  volume    = {15141},
  pages     = {125--140},
  address   = {Milan, Italy},
  publisher = {Springer},
  doi       = {10.1007/978-3-031-73010-8_8},
  url       = {https://doi.org/10.1007/978-3-031-73010-8_8},
  year      = {2024}
}

@inproceedings{huo2025sid,
  title     = {Self-Introspective Decoding: Alleviating Hallucinations for
               Large Vision-Language Models},
  author    = {Huo, Fushuo and Xu, Wenchao and Zhang, Zhong and Wang, Haozhao
               and Chen, Zhicheng and Zhao, Peilin},
  booktitle = {The Thirteenth International Conference on Learning
               Representations},
  url       = {https://openreview.net/forum?id=rsZwwjYHuD},
  year      = {2025}
}

@inproceedings{favero2024m3id,
  title     = {Multi-Modal Hallucination Control by Visual Information
               Grounding},
  author    = {Favero, Alessandro and Zancato, Luca and Trager, Matthew and
               Choudhary, Siddharth and Perera, Pramuditha and Achille,
               Alessandro and Swaminathan, Ashwin and Soatto, Stefano},
  booktitle = {Proceedings of the IEEE/CVF Conference on Computer Vision and
               Pattern Recognition (CVPR)},
  pages     = {14303--14312},
  month     = {June},
  address   = {Seattle, WA, USA},
  publisher = {IEEE},
  doi       = {10.1109/CVPR52733.2024.01356},
  url       = {https://openaccess.thecvf.com/content/CVPR2024/html/Favero_Multi-Modal_Hallucination_Control_by_Visual_Information_Grounding_CVPR_2024_paper.html},
  year      = {2024}
}

@inproceedings{zou2025memvr,
  title     = {Look Twice Before You Answer: Memory-Space Visual Retracing for
               Hallucination Mitigation in Multimodal Large Language Models},
  author    = {Zou, Xin and Wang, Yizhou and Yan, Yibo and Lyu, Yuanhuiyi and
               Zheng, Kening and Huang, Sirui and Chen, Junkai and Jiang,
               Peijie and Liu, Jia and Tang, Chang and Hu, Xuming},
  booktitle = {Proceedings of the 42nd International Conference on Machine
               Learning},
  series    = {Proceedings of Machine Learning Research},
  volume    = {267},
  pages     = {80873--80899},
  address   = {Vancouver, Canada},
  publisher = {PMLR},
  url       = {https://proceedings.mlr.press/v267/zou25e.html},
  year      = {2025}
}

@inproceedings{cho-kim-2026-revisit,
  title     = {Revisit What You See: Revealing Visual Semantics in Vision
               Tokens to Guide {LVLM} Decoding},
  author    = {Cho, Beomsik and Kim, Jaehyung},
  booktitle = {Proceedings of the 64th Annual Meeting of the Association for
               Computational Linguistics (Volume 1: Long Papers)},
  pages     = {5794--5824},
  month     = {July},
  address   = {San Diego, California, United States},
  publisher = {Association for Computational Linguistics},
  doi       = {10.18653/v1/2026.acl-long.262},
  url       = {https://aclanthology.org/2026.acl-long.262/},
  year      = {2026}
}

@inproceedings{jiang2025interpreting,
  title     = {Interpreting and Editing Vision-Language Representations to
               Mitigate Hallucinations},
  author    = {Jiang, Nicholas and Kachinthaya, Anish and Petryk, Suzanne and
               Gandelsman, Yossi},
  booktitle = {The Thirteenth International Conference on Learning
               Representations},
  url       = {https://openreview.net/forum?id=94kQgWXojH},
  year      = {2025}
}

@inproceedings{li2023pope,
  title     = {Evaluating Object Hallucination in Large Vision-Language Models},
  author    = {Li, Yifan and Du, Yifan and Zhou, Kun and Wang, Jinpeng and
               Zhao, Xin and Wen, Ji-Rong},
  booktitle = {Proceedings of the 2023 Conference on Empirical Methods in
               Natural Language Processing},
  pages     = {292--305},
  month     = {December},
  address   = {Singapore},
  publisher = {Association for Computational Linguistics},
  doi       = {10.18653/v1/2023.emnlp-main.20},
  url       = {https://aclanthology.org/2023.emnlp-main.20/},
  year      = {2023}
}

@inproceedings{chen2024halc,
  title     = {{HALC}: Object Hallucination Reduction via Adaptive
               Focal-Contrast Decoding},
  author    = {Chen, Zhaorun and Zhao, Zhuokai and Luo, Hongyin and Yao, Huaxiu
               and Li, Bo and Zhou, Jiawei},
  booktitle = {Proceedings of the 41st International Conference on Machine
               Learning},
  series    = {Proceedings of Machine Learning Research},
  volume    = {235},
  pages     = {7824--7846},
  address   = {Vienna, Austria},
  publisher = {PMLR},
  url       = {https://proceedings.mlr.press/v235/chen24bi.html},
  year      = {2024}
}

@inproceedings{wang2025deco,
  title     = {{MLLM} can see? Dynamic Correction Decoding for Hallucination
               Mitigation},
  author    = {Wang, Chenxi and Chen, Xiang and Zhang, Ningyu and Tian, Bozhong
               and Xu, Haoming and Deng, Shumin and Chen, Huajun},
  booktitle = {The Thirteenth International Conference on Learning
               Representations},
  url       = {https://openreview.net/forum?id=4z3IguA4Zg},
  year      = {2025}
}

@inproceedings{su2025activation,
  title     = {Activation Steering Decoding: Mitigating Hallucination in Large
               Vision-Language Models through Bidirectional Hidden State
               Intervention},
  author    = {Su, Jingran and Chen, Jingfan and Li, Hongxin and Chen, Yuntao
               and Qing, Li and Zhang, Zhaoxiang},
  booktitle = {Proceedings of the 63rd Annual Meeting of the Association for
               Computational Linguistics (Volume 1: Long Papers)},
  pages     = {12964--12974},
  month     = {July},
  address   = {Vienna, Austria},
  publisher = {Association for Computational Linguistics},
  doi       = {10.18653/v1/2025.acl-long.634},
  url       = {https://aclanthology.org/2025.acl-long.634/},
  year      = {2025}
}

@inproceedings{wang2024icd,
  title     = {Mitigating Hallucinations in Large Vision-Language Models with
               Instruction Contrastive Decoding},
  author    = {Wang, Xintong and Pan, Jingheng and Ding, Liang and Biemann,
               Chris},
  booktitle = {Findings of the Association for Computational Linguistics: ACL
               2024},
  pages     = {15840--15853},
  month     = {August},
  address   = {Bangkok, Thailand},
  publisher = {Association for Computational Linguistics},
  doi       = {10.18653/v1/2024.findings-acl.937},
  url       = {https://aclanthology.org/2024.findings-acl.937/},
  year      = {2024}
}

@inproceedings{kim2024code,
  title     = {{CODE}: Contrasting Self-generated Description to Combat
               Hallucination in Large Multi-modal Models},
  author    = {Kim, Junho and Kim, Hyun Jun and Kim, Yeon Ju and Ro, Yong Man},
  booktitle = {Advances in Neural Information Processing Systems},
  editor    = {Globerson, Amir and Mackey, Lester and Belgrave, Danielle and
               Fan, Angela and Paquet, Ulrich and Tomczak, Jakub and Zhang,
               Cheng},
  volume    = {37},
  pages     = {133571--133599},
  publisher = {Curran Associates, Inc.},
  doi       = {10.52202/079017-4246},
  url       = {https://proceedings.neurips.cc/paper_files/paper/2024/file/f1592b0d4ab737e18bb1899484d28d96-Paper-Conference.pdf},
  year      = {2024}
}

@inproceedings{gong2024damro,
  title     = {{DAMRO}: Dive into the Attention Mechanism of {LVLM} to Reduce
               Object Hallucination},
  author    = {Gong, Xuan and Ming, Tianshi and Wang, Xinpeng and Wei, Zhihua},
  booktitle = {Proceedings of the 2024 Conference on Empirical Methods in
               Natural Language Processing},
  pages     = {7696--7712},
  month     = {November},
  address   = {Miami, Florida, USA},
  publisher = {Association for Computational Linguistics},
  doi       = {10.18653/v1/2024.emnlp-main.439},
  url       = {https://aclanthology.org/2024.emnlp-main.439/},
  year      = {2024}
}

@misc{nostalgebraist2020logitlens,
  title     = {interpreting {GPT}: the logit lens},
  author    = {nostalgebraist},
  howpublished = {Blog post, {LessWrong}},
  month     = {August},
  note      = {Published 31 August 2020; cross-posted to the {AI} {Alignment}
               {Forum}. Accessed 2026-08-04},
  url       = {https://www.lesswrong.com/posts/AcKRB8wDpdaN6v6ru/interpreting-gpt-the-logit-lens},
  year      = {2020}
}

@misc{liu2024llavanext,
  title     = {{LLaVA-NeXT}: Improved reasoning, {OCR}, and world knowledge},
  author    = {Liu, Haotian and Li, Chunyuan and Li, Yuheng and Li, Bo and
               Zhang, Yuanhan and Shen, Sheng and Lee, Yong Jae},
  howpublished = {Blog post, {LLaVA} project blog},
  month     = {January},
  note      = {Published 30 January 2024. Accessed 2026-08-04},
  url       = {https://llava-vl.github.io/blog/2024-01-30-llava-next/},
  year      = {2024}
}

\appendix
\newpage
\clearpage

\section{Experimental and Evaluation Details}
\label{sec:implementation-details}

\subsection{Model Checkpoints}

Table~\ref{tab:model-checkpoints} lists the exact model checkpoints used in our
experiments together with their language backbones.  We use the abbreviated
model names in the first column consistently throughout the paper.

\begin{table*}[t]
    \centering
    \small
    \begin{tabular}{lll}
        \toprule
        Model & Checkpoint ID & Language backbone \\
        \midrule
        \instructblipseven
        & \texttt{Salesforce/LAVIS blip2\_vicuna\_instruct/vicuna7b}
        & Vicuna-7B \\
        \llavaonefive
        & \texttt{liuhaotian/llava-v1.5-7b} & Vicuna-7B-v1.5 \\
        \llavanextseven
        & \texttt{liuhaotian/llava-v1.6-mistral-7b}
        & Mistral-7B-Instruct-v0.2 \\
        \qwenvlseven
        & \texttt{Qwen/Qwen2.5-VL-7B-Instruct} & Qwen2.5-7B \\
        InternVL2.5-8B
        & \texttt{OpenGVLab/InternVL2\_5-8B}
        & \texttt{internlm/internlm2\_5-7b-chat} \\
        Gemma-3-12B
        & \texttt{google/gemma-3-12b-it} & Gemma-3-12B \\
        InstructBLIP-13B
        & \texttt{Salesforce/LAVIS blip2\_vicuna\_instruct/vicuna13b}
        & \texttt{lmsys/vicuna-13b-v1.1} \\
        LLaVA-1.5-13B
        & \texttt{liuhaotian/llava-v1.5-13b} & Vicuna-13B-v1.5 \\
        InternVL2.5-26B
        & \texttt{OpenGVLab/InternVL2\_5-26B}
        & \texttt{internlm/internlm2\_5-20b-chat} \\
        Gemma-3-27B
        & \texttt{google/gemma-3-27b-it} & Gemma-3-27B \\
        Qwen2.5-VL-32B
        & \texttt{Qwen/Qwen2.5-VL-32B-Instruct} & Qwen2.5-32B \\
        \bottomrule
    \end{tabular}
    \caption{\textbf{Model checkpoints and their language backbones.}}
    \label{tab:model-checkpoints}
\end{table*}

\subsection{Calibration Data and Operating-Point Selection}

For each backbone, we sample 500 images from MS~COCO train2014 \citep{lin2014coco} with seed 4242.
The calibration images are disjoint from both the seed-42 tuning subset used
for operating-point selection and the evaluation manifests.  We run the
unmodified model greedily and, at each step, collect the DMRR value of every token in
the shortest probability prefix whose mass reaches 0.9.  We enforce a minimum
of two candidates and cap the prefix at 50 tokens.  We condition token
observations on candidacy and do not sweep the vocabulary at every decoding
step.

\begin{table*}[t]
    \centering
    \small
    \setlength{\tabcolsep}{4pt}
    \begin{tabular}{lrrrr}
        \toprule
        Backbone & $\ell$ & $\beta$ & $b_0$ & Registered \\
        \midrule
        LLaVA-1.5   & 29 & 1.1 & 0.00324809  & 2,580 \\
        LLaVA-NeXT  & 30 & 1.2 & 0.00274012  & 3,734 \\
        InstructBLIP & 29 & 0.5 & 0.00184383 & 2,708 \\
        Qwen2.5-VL  & 27 & 1.1 & 0.000653155 & 7,028 \\
        \bottomrule
    \end{tabular}
    \caption{\textbf{Frozen backbone-specific operating points and
    reference-stability table statistics.}}
    \label{tab:operating-points}
\end{table*}

For operating-point selection, we sample 500 images from MS~COCO train2014
with seed 42. The initial grid pairs the three decoder layers with
$\beta\in\{0.5,0.7,0.9,1.1,1.3\}$, which gives 15 configurations. We then
identify the most promising pair of adjacent strengths and evaluate their
midpoint at all three layers. This adds three configurations and gives 18 in
total per backbone. We inspect CHAIR$_S$, CHAIR$_I$, recall, and F1 together
and select the operating point qualitatively rather than optimizing a single
aggregate score. We then freeze the selected layer and $\beta$ for all
remaining benchmarks.
Table~\ref{tab:operating-points} lists the frozen operating point ($\ell$,
$\beta$, $b_0$) and the number of registered tokens for each backbone. We
report CHAIR results on 500 images sampled from MS~COCO val2014. MS~COCO
defines train2014 and val2014 as distinct image splits and explicitly limits
near-duplicates across splits \citep{lin2014coco}. No evaluation image is
therefore used for operating-point selection, and we use no benchmark-specific
strength.

The calibration-size sweep in Figure~\ref{fig:calibration-size} uses a
single nested sequence built from the frozen seed-4242 panel. We sort the
panel by COCO image ID and take the first $n$ images as the subsets up to
$n{=}500$. For larger sizes, we continue the same random stream, sample 500
additional images without replacement from the remaining pool, sort them by
image ID, and append them to the frozen panel. The 700- and 1,000-image sets
are prefixes of this extended sequence, so every smaller subset is contained
in every larger one. The 500-image arm reproduces the deployed configuration
exactly. Refitting from this prefix recovered the deployed $b_0$, all 2,580
registered token references, and their observation counts, and the arm
reuses the frozen evaluation output.

\subsection{Evaluation Protocols}
\label{sec:evaluation-prompts}

Table~\ref{tab:evaluation-inventory} summarizes the evaluation splits and
reported metrics. Unless otherwise stated, we follow the setup in each
benchmark's original paper and released official evaluator as closely as
possible. We select CHAIR hyperparameters on the seed-42 COCO train2014 tuning
subset and report results on the separate COCO val2014 subset listed in the
table. We hold prompts, decoding limits, and evaluator contracts fixed across
the base model, \methodname{}, and the compared methods.

\begin{table*}[t]
    \centering
    \small
    \setlength{\tabcolsep}{4pt}
    \begin{tabular}{llrrl}
        \toprule
        Benchmark & Split & Records & Max. & Reported metrics \\
        \midrule
        CHAIR & MS~COCO val500 & 500 & 512
          & $C_S$, $C_I$, object recall, F1, length \\
        AMBER--Generative & official generative & 1,004 & 128
          & CHAIR, Cover, Hal., Cog. \\
        \multirow{2}{*}{AMBER--Discriminative}
          & \multirow{2}{*}{official discriminative}
          & \multirow{2}{*}{14,216}
          & \multirow{2}{*}{8}
          & accuracy, precision, recall, F1, \\
          & & & & unparsed, AMBER score \\
        MMHal-Bench & full set & 96 & 256
          & quality score, hallucination rate \\
        MM-Vet & original test & 218 & 1,024
          & total score \\
        \bottomrule
    \end{tabular}
    \caption{\textbf{Evaluation datasets, record counts, maximum numbers of
    newly generated tokens (Max.), and reported metrics.}}
    \label{tab:evaluation-inventory}
\end{table*}

\subsubsection{CHAIR}

We evaluate CHAIR \citep{rohrbach2018object} on a fixed 500-image subset of
MS~COCO val2014, disjoint from the tuning and calibration images. The image
identifiers are fixed by the evaluation manifest sampled with seed 42.
We query every image with the exact prompt ``Please describe this image in
detail.'' and cap generation at 512 new tokens. We use the official
CHAIR object vocabulary and synonym mapping. Following the benchmark
protocol, the reference object set for each image is the union of objects in
the MS~COCO instance annotations and the reference captions.

CHAIR$_S$ is the fraction of captions that contain at least one hallucinated
object mention, whereas CHAIR$_I$ is the fraction of generated object mentions
that are hallucinated. Object recall is corpus-micro recall, computed by
summing the numbers of recovered and reference object categories across
images. We additionally derive object precision as $1-\mathrm{CHAIR}_I$ and report
its harmonic mean with object recall as F1. Length is the mean number of
whitespace-delimited words in the generated captions. Every method uses the
same fixed manifest, prompt, synonym mapping, and scorer. As described above,
we select the readout layer $\ell$ and $\beta$ on the COCO train2014 tuning
subset before scoring the 500 COCO val2014 captions.

\subsubsection{AMBER}

AMBER \citep{wang2023amber} contains a 1,004-example generative split and a
14,216-question discriminative split. The latter comprises 4,924 existence,
7,628 attribute, and 1,664 relation questions. For the generative split, we
use the exact prompt ``Describe this image.'' and generate at most 128 new
tokens. The official noun-based scorer extracts and lemmatizes nouns with NLTK
and matches them using the released association and safe-word resources
together with the \texttt{en\_core\_web\_lg} similarity model at threshold
0.8. We report its CHAIR, object coverage (Cover), caption-level hallucination
rate (Hal.), and cognitive hallucination (Cog.) metrics. These quantities use
AMBER's released vocabulary and should not be numerically equated with
MS~COCO CHAIR$_I$.

For the discriminative split, we preserve each released yes/no question
verbatim, add no answer-format suffix, and cap generation at 8 new tokens. To
avoid silently mapping a free-form response to a binary class, we normalize
Unicode and case, remove leading whitespace and punctuation, and accept a
response only if its first token is \textit{yes} or \textit{no}. We mark all other
responses as unparsed and count them as incorrect for accuracy. An unparsed
example whose ground-truth label is \textit{no} also counts as a false
negative for recall. We treat \textit{no}, which denotes the hallucinatory
case, as the positive class and report accuracy, precision, recall, F1, and the
number and percentage of unparsed responses. F1 uses the scorer's $10^{-4}$
denominator smoothing. The reported combined score is
\begin{equation}
    \mathrm{AMBER\ Score}
    = \tfrac{1}{2}\left[(100-\mathrm{CHAIR})+\mathrm{F1}\right],
\end{equation}
with both constituent metrics expressed on a 0--100 scale.

\subsubsection{MMHal-Bench}

We use all 96 MMHal-Bench examples \citep{sun2023factually}, comprising 12
examples from each of eight question types: attribute, adversarial,
comparison, counting, relation, environment, holistic, and other. We pass the
released question verbatim without an additional prompt suffix and allow at
most 256 new tokens. We retain the released five-example evaluator prompt and
rubric without modification. This text-only evaluator receives the released
image-content object list, question, human reference answer, and model
response. It does not receive the image itself. Its rubric assigns an integer
score from 0 to 6 according to informativeness and the presence of
hallucination. Scores 0--2 denote a response with hallucination, and scores
3--6 denote a response without hallucination. We report the mean score and the
percentage of examples assigned a score below 3.

All configurations are judged with the frozen
\texttt{gpt-4.1-mini-2025-04-14} snapshot at temperature 0. The parser extracts
a ``Rating: $s$'' entry for $s\in\{0,\ldots,6\}$, and an output that does not
yield a unique valid rating receives score 0. Thus, parse failures receive
identical treatment for every method and stay in the denominator.

\subsubsection{MM-Vet}

We evaluate the original MM-Vet test set \citep{yu2024mmvet}, which contains
218 questions over 200 images. Each released question is used verbatim without
an answer-format suffix, and generation is capped at 1,024 new tokens. The
official text-only grading contract retains the released six-example prompt
without modification. It compares the question, released reference answer,
and model prediction, and returns one of
$\{0.0,0.1,\ldots,1.0\}$. The prompt explicitly preserves the benchmark's
\texttt{<AND>} semantics, which require all referenced elements, and
\texttt{<OR>} semantics, which accept any listed alternative. We use
\texttt{gpt-4.1-mini-2025-04-14} at temperature 0. An evaluator output that
does not yield a valid score receives score 0.

The total MM-Vet score is 100 times the mean item grade. The evaluator also
defines capability breakdowns over recognition (150), OCR (96), knowledge
(84), generation (80), spatial reasoning (75), and mathematics (26). These
capability tags overlap, so their sample counts do not form a partition of the
218 questions. The same judge snapshot, prompt contract, and one-grade policy
are used for every MM-Vet row in Table~\ref{tab:mmhal-bench}.

\subsection{Generation Settings}

The base model and \methodname{} use batch size 1, eager attention, one beam,
and no sampling.  The generation seed is 42.  The checkpoint-native repetition
penalty is 1.05 for Qwen2.5-VL and 1.0 for the other backbones.
Benchmark-specific decoding limits are fixed in the evaluation manifests, and
the method and base model use the same limit and prompt. For all baseline
methods, we use the authors' official implementations and the default
configurations specified in the corresponding papers and repositories. Each
baseline therefore keeps the decoding regime of its original paper. OPERA
searches with five beams, which also accounts for the decoding cost we report
for it in Table~\ref{tab:chair-amber}, and VCD samples instead of decoding
greedily.

\subsection{Hardware and Software}

Each generation job uses an NVIDIA A100-SXM4-80GB GPU, with approximately
250 total GPU-hours used across all experiments.  LLaVA-1.5 uses
Python~3.10.20, PyTorch~2.1.2+cu121, and Transformers~4.40.0.  LLaVA-NeXT and
Qwen2.5-VL use Python~3.10.20, PyTorch~2.5.1+cu124, and
Transformers~4.52.1.  The original LAVIS InstructBLIP environment uses
Python~3.9.16, PyTorch~2.0.1+cu117, and Transformers~4.33.2.  LLaVA and InstructBLIP run in FP16, and Qwen2.5-VL runs in BF16.
For every run, we record the model revision, the evaluation-manifest and
prompt hashes, the calibration-table hash, the decoding configuration, and the
runtime environment.

\subsection{Artifact Licenses and Intended Use}
\label{sec:artifact-licenses}

We use all datasets, model checkpoints, and evaluation tools for
non-commercial research, consistent with their intended use and stated
terms. MS~COCO annotations are released under the Creative Commons
Attribution 4.0 license, and the underlying images are governed by the
Flickr Terms of Use \citep{lin2014coco}. AMBER \citep{wang2023amber} and
MMHal-Bench \citep{sun2023factually} are released under the Apache-2.0
license, and GQA \citep{hudson2019gqa} under CC~BY~4.0. The MM-Vet
evaluation code is Apache-2.0 and its data are distributed under
CC~BY-NC~4.0 \citep{yu2024mmvet}. Among the checkpoints in
Table~\ref{tab:model-checkpoints}, \qwenvlseven{} and \llavanextseven{}
(Mistral-7B-Instruct-v0.2 backbone) are Apache-2.0, \llavaonefive{} is
subject to the Llama~2 Community License through its Vicuna-v1.5 backbone,
and \instructblipseven{} follows the non-commercial research terms of its
LLaMA-based Vicuna backbone. The AMBER scorer depends on NLTK (Apache-2.0)
and the spaCy \texttt{en\_core\_web\_lg} model (MIT), and the LLM judge is
accessed through the OpenAI API under its terms of use. Qualitative
figures reproduce individual benchmark images solely for scholarly
analysis.

\section{Additional Diagnostic Analyses}
\label{sec:diagnostic-analyses}

\subsection{Scope of the Diagnostic Analyses}

The analyses in Section~\ref{sec:motivation} answer two design questions.
The first asks which property of a visual-position vocabulary readout can be pooled across positions without inheriting position-specific cardinal scale.
The second asks which reference makes the pooled evidence interpretable for a particular token.
These analyses motivate an intervention design.
We do not claim that rank is a universally superior hallucination detector or that token-specific centering improves standalone classification.

We conduct the diagnostic analyses on \llavaonefive{} and \qwenvlseven{} at their selected readout layers.
The scale-dispersion and token-identity-share measurements additionally cover all four evaluated backbones.
The two models have different visual-token counts and different absolute DMRR scales.
We therefore report within-backbone effects and do not compare absolute evidence values across models.

\subsection{Rank-Preserving Scale Analysis}
\label{sec:rank-scale-analysis}

Let $\mathbf z_j^{(\ell)}$ be the vocabulary readout of visual position $j$.
Multiplying this vector by a positive scalar preserves its complete vocabulary ordering but changes the probability distribution obtained by softmax.
The controlled test retains the observed readout directions and reassigns positive scale factors among visual positions.
It is an invariance stress test rather than a model of a naturally occurring perturbation.
Its role is to isolate sensitivity to position-wise cardinal scale while every within-position vocabulary ordering stays fixed.
Consequently, every token retains its rank within each position, while a probability aggregate can change because the positions now contribute distributions with different sharpness.
The probability arm replaces DMRR with the logarithm of each token's position-averaged softmax probability, floored at $10^{-8}$, and mirrors the calibration of Section~\ref{sec:calib}: the token reference is the candidate-conditioned median of this readout, and registration applies the same $\Delta e(v)<0.5$ rule.
One substitution is forced by the readout's range.
The pooled median of the log-probability readout is negative ($-8.6$ for \llavaonefive{} and $-10.2$ for \qwenvlseven{}), so reusing it as the normalization scale $b_0$ in Eq.~\eqref{eq:normalized-edit} would flip the sign of the normalized deficit, clipping every evidence deficit to zero while directing suppression toward evidence surpluses.
The probability arm therefore normalizes by a positive scale solved on the calibration images so that its mean deployed edit matches the DMRR arm's under the same maximum penalty ($0.966$ for \llavaonefive{} and $0.905$ for \qwenvlseven{}), and the registration rule is evaluated under the same positive scale.
With this substitution, the two arms share the bounded evidence-deficit form of Eqs.~\eqref{eq:normalized-edit} and~\eqref{eq:suppression-strength} under a matched mean edit budget and differ in the underlying readout and its normalization scale.

The maximum rank change under this manipulation is zero.
With the original calibration table fixed, the probability-derived suppression strengths shift by a mean absolute $0.0445$ for LLaVA and $0.0462$ for Qwen, whereas the rank-derived suppression strengths remain exactly unchanged.
Equalizing the scale closes only $6.94\%$ and $8.76\%$ of the full probability-to-rank gap in suppression strengths.
The equalization comparison uses the raw position-averaged probability rather than its logarithm, whose pooled calibration median is positive and therefore serves directly as the normalization scale, with the penalty refit so this arm meets the same mean edit budget as the DMRR arm.
Scale equalization also leaves most of the fixed-rank dispersion in place.
Setting every position's logit scale to the image median reduces the rank-10 p90/p10 spread from $13.82\times$ to $11.05\times$ on LLaVA and from $7.79\times$ to $4.25\times$ on Qwen.
Between $67\%$ and $95\%$ of the log-probability spread survives across the measured depths 1--1,000 for both backbones (Table~\ref{tab:depth-dispersion}).
Position-wise distribution shape therefore differs beyond what a per-position rescaling can correct.
The result identifies cardinal scale as one nuisance in probability pooling, not as a complete explanation of the end-to-end difference between the two interventions.

Three tables document the position-scale diagnostics.
Table~\ref{tab:scale-dispersion} extends the dispersion measurement to all four evaluated backbones.
It fixes a probability threshold and reports the across-position spread of its rank depth, complementing Figure~\ref{fig:scale-nuisance}, which fixes a vocabulary rank and reports the spread of its probability.
At every selected readout layer, the rank depth of a fixed probability threshold varies across positions by at least a factor of four on 5,000 development images, and the 200-image measurement closely matches.
Table~\ref{tab:depth-dispersion} reports the depth-wise spreads on 500 development images before and after scale equalization.
Its rank-10 rows underlie Figure~\ref{fig:scale-nuisance}, and the original rank-10 spreads carry retrospective image-bootstrap 95\% intervals of $[2.568,\,2.723]$ and $[2.032,\,2.068]$ in log units, or $[13.0\times,\,15.2\times]$ and $[7.6\times,\,7.9\times]$ as ratios.
Table~\ref{tab:rescale-invariance} reports the rescaling stress test and the scale-equalization closure with their bootstrap intervals.

\begin{table*}[t]
    \centering
    \small
    \setlength{\tabcolsep}{6pt}
    \begin{tabular}{lrrrr}
        \toprule
        Threshold $\tau$ & \llavaonefive & \instructblipseven & \llavanextseven & \qwenvlseven \\
        \midrule
        $10^{-3}$ & 8.6 \;(8.5)   & 4.1 \;(3.9) & 9.2 \;(9.2)     & 5.8 \;(5.7) \\
        $10^{-4}$ & 62.5 \;(59.4) & 7.1 \;(6.7) & 87.7 \;(84.0)   & 7.4 \;(7.4) \\
        $10^{-5}$ & 93.8 \;(89.4) & 4.6 \;(4.6) & 477.3 \;(459.4) & 11.5 \;(10.9) \\
        \bottomrule
    \end{tabular}
    \caption{\textbf{Across-position dispersion of the rank depth of a fixed
    probability threshold.} Each cell is the image-median p90/p10 ratio of
    $N_j(\tau)$, the number of vocabulary items with probability at least
    $\tau$ at visual position $j$. Values in parentheses use the 200-image
    measurement.}
    \label{tab:scale-dispersion}
\end{table*}

\begin{table*}[t]
    \centering
    \small
    \setlength{\tabcolsep}{6pt}
    \begin{tabular}{r rrr rrr}
        \toprule
        & \multicolumn{3}{c}{\llavaonefive} & \multicolumn{3}{c}{\qwenvlseven} \\
        \cmidrule(lr){2-4}\cmidrule(lr){5-7}
        Rank depth $k$ & Original & Equalized & Surv. & Original & Equalized & Surv. \\
        \midrule
        1     & 4.64 & 3.86 & 83\% & 4.11 & 3.53 & 86\% \\
        2     & 3.59 & 2.91 & 81\% & 3.29 & 2.70 & 82\% \\
        5     & 2.55 & 2.15 & 84\% & 2.47 & 1.87 & 76\% \\
        10    & 2.63 & 2.40 & 91\% & 2.05 & 1.45 & 70\% \\
        20    & ---  & ---  & ---  & 1.79 & 1.20 & 67\% \\
        50    & 3.43 & 3.26 & 95\% & 1.58 & 1.18 & 75\% \\
        100   & 3.87 & 3.64 & 94\% & 1.53 & 1.28 & 84\% \\
        500   & 4.63 & 4.14 & 90\% & ---  & ---  & --- \\
        1,000 & 4.84 & 4.22 & 87\% & 1.83 & 1.57 & 86\% \\
        \bottomrule
    \end{tabular}
    \caption{\textbf{Across-position dispersion of the rank-$k$ log
    probability before and after scale equalization.} Each cell is the
    image-median p90--p10 spread in log units. Surv.\ is the share of the
    original spread that survives equalization. Depths absent from a
    backbone's stored grid are marked ---.}
    \label{tab:depth-dispersion}
\end{table*}

\begin{table*}[t]
    \centering
    \small
    \setlength{\tabcolsep}{5pt}
    \begin{tabular}{llrlrl}
        \toprule
        & & \multicolumn{2}{c}{\llavaonefive} & \multicolumn{2}{c}{\qwenvlseven} \\
        \cmidrule(lr){3-4}\cmidrule(lr){5-6}
        Signal & Quantity & Value & 95\% CI & Value & 95\% CI \\
        \midrule
        \rowcolor{gray!15}
        Rank & Suppression-strength change & $0^{\dagger}$ & --- & $0^{\dagger}$ & --- \\
        \addlinespace
        Probability & Suppression-strength mean $|\Delta|$ & 0.0445 & $[0.0430,\,0.0460]$ & 0.0462 & $[0.0447,\,0.0478]$ \\
             & Candidate-score Spearman & 0.964 & --- & 0.961 & --- \\
             & Edited-set Jaccard & 0.805 & --- & 0.857 & --- \\
             & Strongest edited candidate changed & 18.6\% & --- & 17.8\% & --- \\
             & One-step winner changed & 1.7\% & --- & 2.5\% & --- \\
        \addlinespace
        Equalization & Probability-to-rank gap closed & 6.94\% & $[6.52,\,7.36]$ & 8.76\% & $[6.50,\,11.11]$ \\
        \bottomrule
    \end{tabular}
    \caption{\textbf{Order-preserving rescaling stress test at the selected
    readout layers.} The probability rows report how the probability-derived
    edit field reacts to the rescaling, and the winner is the candidate-local
    one-step argmax after the edit. The equalization row reports the closed
    share of the probability-to-rank gap. Intervals are image-bootstrap
    95\%, retrospective for the rescaling rows and paired for the
    equalization row.
    $\dagger$ marks an exact zero. The shaded row is the rank signal used by
    \methodname{}.}
    \label{tab:rescale-invariance}
\end{table*}

\begin{table*}[t]
    \centering
    \small
    \setlength{\tabcolsep}{3.5pt}
    \begin{tabular}{llrrrrrrrr}
        \toprule
        & & \multicolumn{4}{c}{\llavaonefive} & \multicolumn{4}{c}{\qwenvlseven} \\
        \cmidrule(lr){3-6}\cmidrule(lr){7-10}
        Support & Readout policy & CHAIR$_S\downarrow$ & CHAIR$_I\downarrow$ & Recall$\uparrow$ & F1$\uparrow$
        & CHAIR$_S\downarrow$ & CHAIR$_I\downarrow$ & Recall$\uparrow$ & F1$\uparrow$ \\
        \midrule
        \rowcolor{gray!15}
        Matched & Ordinal readout (DMRR) & 45.8 & 12.78 & 75.2 & 80.77 & 26.0 & 8.64 & 58.0 & 70.99 \\
                & Standardized ($z$-score) & 52.8 & 14.87 & 76.9 & 80.83 & 24.4 & 6.45 & 55.6 & 69.77 \\
        \addlinespace
        Native  & Base decoding & 52.6 & 16.07 & 77.3 & 80.48 & 28.8 & 8.26 & 56.6 & 69.97 \\
                \rowcolor{gray!15}
                & Ordinal readout (DMRR) & 45.8 & 12.77 & 75.2 & 80.77 & 25.8 & 8.59 & 57.8 & 70.82 \\
                & Standardized ($z$-score) & 52.4 & 14.90 & 76.9 & 80.81 & 24.0 & 6.38 & 55.0 & 69.29 \\
                & Mean probability & 49.2 & 13.30 & 77.4 & 81.81 & 25.2 & 8.94 & 57.6 & 70.58 \\
                & Standardized probability & 49.4 & 13.80 & 77.0 & 81.34 & 26.6 & 7.17 & 57.3 & 70.86 \\
        \bottomrule
    \end{tabular}
    \caption{\textbf{End-to-end control between the ordinal and standardized
    readouts.} Matched rows evaluate both policies on identical candidate
    support. Native rows let each policy form its own candidate sets, which is
    why the two blocks differ slightly. The native probability rows are not
    the matched-budget rank--probability comparison of
    Table~\ref{tab:detection-not-mitigation}. Shaded rows are the ordinal
    readout used by \methodname{}. Arrows indicate the preferred direction.}
    \label{tab:normalized-readout-control}
\end{table*}

Table~\ref{tab:normalized-readout-control} reports the standardized-readout control referenced in Section~\ref{sec:scale}.
The control replaces the ordinal readout with a per-position standardized ($z$-scored) readout of the same lens under matched calibration and suppression.
The comparison runs in a self-contained harness in which each backbone regenerates its own base, with \qwenvlseven{} at repetition penalty 1.0, so its values are not numerically comparable with the main results.
On \llavaonefive{}, the ordinal readout lowers CHAIR$_S$ from $52.8$ to
$45.8$ under matched support, while recall decreases from $76.9$ to $75.2$.
On \qwenvlseven{}, the standardized readout lowers CHAIR$_S$ from $26.0$ to
$24.4$, while the ordinal readout raises recall from $55.6$ to $58.0$.
These mixed point estimates do not establish cross-backbone superiority, but
they show that per-position standardization does not reproduce the ordinal
intervention's behavior.

\subsection{Token-Conditioned Reference Analysis}
\label{sec:token-reference-analysis}

After replacing probability magnitude with rank, the remaining evidence still depends on token identity.
In the image-balanced LLaVA decomposition reported in the main paper, $66.36\%$ of DMRR variability is associated with differences between tokens and $33.64\%$ with differences between images for the same token.
The decomposition is descriptive.
Its role is to test whether a single global reference is well matched to the evidence being interpreted.
Each observation is the image-level value $\dmrr_I(v)$ attached to one candidate occurrence of token $v$.
Repeated occurrences within an image share one value because we compute $\dmrr_I(v)$ once per image.
The decomposition splits the total sum of squares of these observations into a between-token component, the variability of per-token means, and a between-image, within-token remainder, and reports each component's share of the total.
The occurrence-weighted variant counts every candidate occurrence once, whereas the image-balanced variant counts every token--image pair once, subject to the rare-token filter stated in Figure~\ref{fig:token-decomposition}.

We evaluate this question out of sample using 500 calibration images and 4,969 disjoint images.
For each token observed in the held-out set, we form the median of its held-out DMRR observations and score a stored reference $b$ by the normalized gap between the two in $b_0$ units.
The token-unweighted aggregate takes the median of this gap with every token counted once, and the candidate-mass-weighted aggregate weights each token by its held-out candidate mass.
The correct token-specific median reduces error relative to a global reference by more than $90\%$ in both analyzed backbones under token-unweighted and candidate-mass-weighted aggregation.
Randomly shuffling the mapping between tokens and reference values preserves the marginal reference distribution but performs worse than the correct mapping in all 1,000 permutations, which corresponds to the smallest attainable one-sided permutation $p$-value under this null ($p=1/1001$).
This control separates the value distribution of the calibration table from the token-to-reference correspondence.

\begin{table*}[t]
    \centering
    \small
    \setlength{\tabcolsep}{6pt}
    \begin{tabular}{lrr}
        \toprule
        \multicolumn{3}{l}{\textit{(a) Weighting sensitivity of the decomposition (\llavaonefive)}} \\
        Analysis unit & Between tokens & Between images, within token \\
        \midrule
        Token--image balanced, min.\ 1 image/token  & 66.40\% & 33.60\% \\
        Token--image balanced, min.\ 2 images/token & 66.33\% & 33.67\% \\
        Token--image balanced, min.\ 5 images/token (primary) & 66.36\% & 33.64\% \\
        Token--image balanced, min.\ 10 images/token & 66.59\% & 33.41\% \\
        Occurrence weighted & 74.62\% & 25.38\% \\
        \midrule
        \addlinespace
        \multicolumn{3}{l}{\textit{(b) Token-identity share (ICC) on 5,000 development images}} \\
        Backbone & ICC (simple) & ICC (ANOVA) \\
        \midrule
        \llavaonefive    & 0.470 & 0.655 \\
        \instructblipseven & 0.440 & 0.455 \\
        \llavanextseven  & 0.441 & 0.456 \\
        \qwenvlseven     & 0.525 & 0.757 \\
        \bottomrule
    \end{tabular}
    \caption{\textbf{Token-identity component of DMRR variability.}
    Panel~(a) varies the rare-token filter and the weighting unit of the
    decomposition in Figure~\ref{fig:token-decomposition}. Panel~(b) reports
    the intraclass correlation of token identity on image-unique observations
    at the selected readout layers.}
    \label{tab:variance-decomposition}
\end{table*}

\begin{table*}[t]
    \centering
    \small
    \setlength{\tabcolsep}{5pt}
    \begin{tabular}{llrr>{\columncolor{gray!15}}rr}
        \toprule
        & & \multicolumn{3}{c}{Out-of-sample error ($b_0$ units)} & \\
        \cmidrule(lr){3-5}
        Backbone & Aggregation & Global & Shuffled & Token-specific & Reduction \\
        \midrule
        \llavaonefive & token-unweighted median & 0.849 & 1.010 & 0.067 & 92.1\% \\
        & candidate-mass-weighted mean & 4.211 & 5.195 & 0.341 & 91.9\% \\
        \addlinespace
        \qwenvlseven & token-unweighted median & 0.831 & 0.510 & 0.077 & 90.7\% \\
        & candidate-mass-weighted mean & 7.506 & 8.501 & 0.612 & 91.9\% \\
        \bottomrule
    \end{tabular}
    \caption{\textbf{Out-of-sample calibration error of global, shuffled,
    and token-specific references.} The shuffled column is the median of 1,000 shuffles, and Reduction is the
    error decrease of the correct reference relative to the global reference.
    The shaded column is the reference used by \methodname{}.}
    \label{tab:token-reference-correspondence}
\end{table*}

Two tables document the token-reference analysis.
Table~\ref{tab:variance-decomposition} varies the rare-token filter and the weighting unit of the decomposition, and it reports the token-identity share for all four evaluated backbones on 5,000 development images.
Table~\ref{tab:token-reference-correspondence} lists the out-of-sample calibration errors behind Figure~\ref{fig:token-correspondence}, including the shuffled-reference null for both analyzed backbones.

\subsection{Detection Quality Does Not Predict Mitigation}
\label{sec:detection-not-mitigation}

This section tests whether standalone detection quality identifies the signal that best drives an autoregressive intervention.
We choose the ordinal readout in Section~\ref{sec:scale} for comparability across visual positions and for calibratability, not for detection quality.
We compare rank- and probability-derived signals at the emitted-object detection level, in a raw score-to-edit negative control, and in the complete calibrated intervention.

\begin{table*}[t]
    \centering
    \small
    \setlength{\tabcolsep}{6pt}
    \begin{tabular}{lrrrrr}
        \toprule
        Signal & CHAIR$_S\downarrow$ & CHAIR$_I\downarrow$ & Removed$\uparrow$ & Introduced$\downarrow$ & Net$\uparrow$ \\
        \midrule
        \rowcolor{gray!15}
        Rank        & 44.8 & 12.67 & 329 & 198 & 131 \\
        Probability & 50.0 & 14.03 & 283 & 227 & 56 \\
        \bottomrule
    \end{tabular}
    \caption{\textbf{Matched-budget end-to-end comparison of the rank and
    probability signals on \llavaonefive{}.} Removed and Introduced count
    hallucinated-object occurrences relative to the same base captions, and
    Net is their difference. The shaded row is the rank signal used by
    \methodname{}. Arrows indicate the preferred direction.}
    \label{tab:detection-not-mitigation}
\end{table*}

At the detection level, the emitted-object hallucination AUROC reaches $0.679$ with the rank readout and $0.718$ with the probability readout on \llavaonefive{}, and $0.619$ and $0.634$ on \qwenvlseven{}.
This metric orders grounded and hallucinated object tokens that the base trajectory emitted.
It does not specify the full candidate-vector edit or the autoregressive trajectory after an intervention.

The raw score-to-edit control asks whether the uncalibrated signals already differ as local intervention targets.
At every step, each arm suppresses exactly one candidate by the same amount, without $b(v)$, registration, normalization, or clipping.
The contrast measures local switch selectivity within a fixed four-token precursor window.
Raw rank minus probability is $-0.49$ pp ($95\%$ image-bootstrap interval $[-1.19,\,+0.22]$) on \llavaonefive{} and $-0.93$ pp ($[-1.72,\,-0.16]$) on \qwenvlseven{}, so the bare rank signal provides no local advantage.
Rank alone therefore does not explain the final method's behavior.

Table~\ref{tab:detection-not-mitigation} gives the corresponding end-to-end comparison on \llavaonefive{} under a matched mean edit budget.
The calibrated rank intervention lowers CHAIR$_S$ from $50.0$ to $44.8$ relative to the calibrated probability intervention.
It also removes more hallucinated-object occurrences from the shared base captions and introduces fewer new ones.
Figure~\ref{fig:rank-probability-case} illustrates the same distinction on one image.
The probability intervention generates an unsupported fork, whereas the rank intervention avoids it while preserving the grounded pizza and dining table.
The example is illustrative.
The aggregate comparison in Table~\ref{tab:detection-not-mitigation}, rather than this selected image, supports the conclusion.

\newcommand{\casehalluc}[1]{\textcolor{red!70!black}{\textbf{#1}}}
\newcommand{\casegrounded}[1]{\textcolor{blue!70!black}{\textbf{#1}}}
\newcommand{\caseheader}[1]{%
    \noindent\colorbox{black!7}{%
        \parbox{\dimexpr\linewidth-2\fboxsep\relax}{\textbf{#1}}}}
\newcommand{\caseanswerbox}[3]{%
    \noindent\colorbox{#1}{%
        \parbox{\dimexpr\linewidth-2\fboxsep\relax}{%
            \textbf{#2}\quad #3}}}

\begin{figure*}[t]
    \centering
    \caseheader{\llavaonefive}
    \vspace{5pt}

    \begin{minipage}[t]{0.24\textwidth}
        \vspace{0pt}
        \centering
        \includegraphics[width=\linewidth]{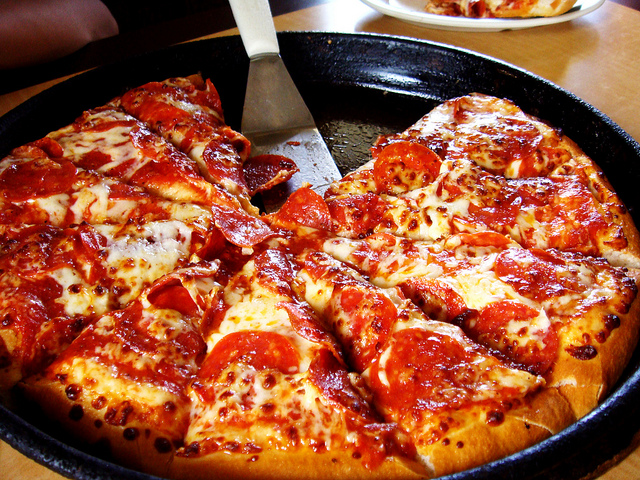}
    \end{minipage}\hfill
    \begin{minipage}[t]{0.73\textwidth}
        \vspace{0pt}
        \small
        \caseanswerbox{red!3}{Base}{The image features a delicious pepperoni pizza with a
        generous amount of pepperoni toppings, placed on a black tray. The pizza
        is cut into slices, making it easy to serve and enjoy. The pizza is
        served on a dining table, and a \casehalluc{fork} is placed nearby, ready for
        use. The table also has a \casehalluc{bowl}, possibly containing additional
        toppings or condiments. The scene is set in a cozy dining area, inviting
        \casehalluc{people} to gather around and enjoy the pizza together.}

        \vspace{5pt}
        \caseanswerbox{yellow!5}{Matched probability}{The image features a delicious
        pepperoni pizza sitting on top of a black pan, ready to be served. The
        pizza is cut into slices and has a generous amount of pepperoni
        toppings. The pan is placed on a dining table, and a \casehalluc{fork} is
        resting nearby, indicating that it is ready to be enjoyed. The pizza
        appears to be freshly baked and is served on a pizza pan, making it an
        appetizing sight for anyone who loves pepperoni pizza.}

        \vspace{5pt}
        \caseanswerbox{blue!3}{$+$\methodname{} (rank)}{The image features a delicious
        pepperoni pizza sitting in a black \casegrounded{pan} on a dining table. The
        pizza is cut into slices, making it easy to serve and enjoy. The pan is
        placed on a dining table, which occupies most of the background of the
        image. The pizza appears to be freshly baked and ready to be eaten.}
    \end{minipage}
    \caption{\textbf{Matched-budget rank and probability interventions on
    one CHAIR image.} The visible utensil is a pizza server rather than a
    fork. Red marks scorer-flagged object
    mentions, and blue marks scorer-matched objects.}
    \label{fig:rank-probability-case}
\end{figure*}

\let\casehalluc\relax
\let\casegrounded\relax
\let\caseheader\relax
\let\caseanswerbox\relax

These results do not isolate a single causal mediator or show that calibration alone creates the gap.
They establish the conclusion relevant to our design.
The ordering of the two signals on standalone detection does not match their ordering on end-to-end mitigation, where the token-calibrated rank intervention performs better under the matched budget.
Detection quality alone therefore does not identify the signal of an effective autoregressive intervention.

\subsection{From Detection to Allocation}
\label{sec:allocation-mechanism}

Table~\ref{tab:allocation-mechanism} examines where the calibrated rank and probability edit fields place their edits on 4,969 held-out images, complementing the raw score-to-edit control of Appendix~\ref{sec:detection-not-mitigation}.
The unit of analysis is the four-step precursor window before each CHAIR object mention.
A window counts as switched when the applied edit changes the one-step argmax at any of its steps, and switch selectivity is the switch rate before hallucinated mentions minus the switch rate before grounded mentions.
A deterministic tokenizer-piece taxonomy, rather than contextual part-of-speech tagging, classifies precursor tokens into structural words, lexical words, punctuation and boundary pieces, and special fragments.
An allocation-exclusive event is a window that switches under exactly one field's candidate assignment, read at that field's first switched step.
The Effect rows exchange only the candidate assignment between the two calibrated fields while the edit-magnitude multiset stays fixed, using either field as the magnitude donor.
On \llavaonefive{}, the rank-derived assignment yields higher hallucination-linked precursor switch selectivity under both donors, while on \qwenvlseven{} the contrasts are directionally consistent but their intervals include zero.
The Role rows decompose the rank-donor effect by precursor token role, and the advantage concentrates on structural tokens.
The Share rows report the fraction of allocation-exclusive first edit events that fall on structural tokens, which is likewise larger under the rank field.
These are descriptive allocation statistics.
They locate the difference between the two calibrated fields without establishing that structural-token edits are individually necessary.

\begin{table*}[t]
    \centering
    \small
    \setlength{\tabcolsep}{5pt}
    \begin{tabular}{llrlrl}
        \toprule
        & & \multicolumn{2}{c}{\llavaonefive} & \multicolumn{2}{c}{\qwenvlseven} \\
        \cmidrule(lr){3-4}\cmidrule(lr){5-6}
        Group & Quantity & Value & 95\% CI & Value & 95\% CI \\
        \midrule
        Effect & Rank-magnitude donor & $+2.32$ pp & $[+1.09,\,+3.53]$ & $+0.55$ pp & $[-1.00,\,+2.12]$ \\
               & Probability-magnitude donor & $+2.06$ pp & $[+0.95,\,+3.14]$ & $+1.37$ pp & $[-0.15,\,+2.91]$ \\
        \addlinespace
        Role   & Structural tokens & $+2.36$ pp & $[+1.28,\,+3.48]$ & $+1.03$ pp & $[-0.36,\,+2.46]$ \\
               & Lexical content tokens & $+0.33$ pp & $[-0.39,\,+1.04]$ & $+0.51$ pp & $[-0.65,\,+1.72]$ \\
               & Punctuation tokens & $-0.74$ pp & $[-1.56,\,+0.09]$ & $+0.72$ pp & $[-1.19,\,+2.62]$ \\
        \addlinespace
        Share  & Rank-allocation-only events & 62.6\% & --- & 55.1\% & --- \\
               & Probability-allocation-only events & 50.0\% & --- & 40.6\% & --- \\
        \bottomrule
    \end{tabular}
    \caption{\textbf{Candidate-allocation analysis of the calibrated rank
    and probability edit fields.} Effect and Role values are
    rank-minus-probability selectivity differences in percentage points with
    image-bootstrap 95\% intervals. Share values are fractions of
    allocation-exclusive events that fall on structural tokens. Role and
    Share use the rank-magnitude donor. The definitions are in the
    accompanying text.}
    \label{tab:allocation-mechanism}
\end{table*}

\subsection{Output Confidence and the Evidence Deficit}
\label{sec:additional-analyses}

Table~\ref{tab:confidence-blindness} asks whether output confidence alone
identifies the hallucinated mentions that the evidence signal targets. Among
object-mention onsets in the top quartile of output confidence, $13.9\%$ are
hallucinated on \llavaonefive{} and $7.3\%$ on \qwenvlseven{}, and the
hallucinated onsets carry a positive median evidence deficit while grounded
onsets in the same quartile carry none. The separation is clear on
\llavaonefive{}. On \qwenvlseven{} the direction agrees but the interval
reaches zero. High output confidence therefore does not subsume the
calibrated evidence signal.

\begin{table*}[t]
    \centering
    \small
    \setlength{\tabcolsep}{5pt}
    \begin{tabular}{lrr}
        \toprule
        Quantity & \llavaonefive & \qwenvlseven \\
        \midrule
        Hallucinated onsets in the top-confidence quartile
        & 13.9\% $[10.3,\,17.7]$ & 7.3\% $[4.4,\,10.7]$ \\
        \quad (count) & 52 / 373 & 20 / 274 \\
        Median evidence deficit at those onsets
        & 1.18 & 0.10 \\
        Median evidence deficit, grounded onsets in the same quartile
        & 0.00 & 0.00 \\
        Median difference $[95\%$ CI$]$
        & 1.18 $[0.47,\,5.56]$ & 0.10 $[0.00,\,7.04]$ \\
        \bottomrule
    \end{tabular}
    \caption{\textbf{Object-mention onsets in the top quartile of output
    confidence.} Brackets are image-bootstrap 95\% intervals, and the
    evidence deficit is $\max(b(v)-\dmrr_I(v),0)/b_0$ before clipping.}
    \label{tab:confidence-blindness}
\end{table*}

\section{Method and Implementation Details}
\label{sec:method-details}

\subsection{Architecture-Specific Visual Positions}

The set $P$ always refers to positions that carry visual information inside the language-model sequence.
For LLaVA-1.5 and LLaVA-NeXT, these are the projected vision-encoder patch embeddings inserted in place of the image placeholder \citep{liu2023llava,liu2024improved,liu2024llavanext}.
For InstructBLIP, $P$ contains the projected Q-Former query embeddings supplied to the Vicuna language model \citep{dai2023instructblip}.
For Qwen2.5-VL, $P$ contains the merged visual embeddings inserted into the multimodal sequence \citep{bai2025qwen25vl}.
We exclude text positions and image-boundary control tokens.

At the selected language-model layer $\ell$, we collect $\mathbf h_j^{(\ell)}$ for every $j\in P$.
The readout applies the final output normalization and the language model's vocabulary head, exactly as in Eq.~\eqref{eq:lens}.
We do not train a translator or probe for intermediate layers.
We define all ranks over the native vocabulary of the corresponding language backbone.

\begin{algorithm*}[t]
\caption{\textbf{Calibration and inference with \methodname.}}
\label{alg:method}
\small
\begin{algorithmic}[1]
\Statex \textbf{Calibration stage}
\Require Unlabeled calibration set $\mathcal D_{\mathrm{cal}}$, frozen model
    $M$, readout layer $\ell$
\Ensure Frozen token table $\mathcal T$
\State Initialize an empty evidence multiset $D_v$ for each token $v$
\For{each image--prompt pair $(I,x)\in\mathcal D_{\mathrm{cal}}$}
    \State Compute $\dmrr_I(v)$ for all tokens using Eq.~\eqref{eq:dmrr}
    \State Decode with the frozen model $M$ and collect candidate sets $\Cand_t$
    \State Add $\dmrr_I(v)$ to $D_v$ whenever $v\in\Cand_t$
\EndFor
\State Compute the pooled scale $b_0$ and token references $b(v)$ by
    Eq.~\eqref{eq:baseline}
\For{each token $v$ with $D_v\neq\varnothing$}
    \State Estimate the reference interval
        $[b_{\mathrm{lo}}(v),b_{\mathrm{hi}}(v)]$
    \State Compute $e(b,x)$ and $\Delta e(v)$ using
        Eqs.~\eqref{eq:normalized-edit} and~\eqref{eq:deltae}
    \If{the interval exists and $\Delta e(v)<0.5$}
        \State Register $(v,b(v))$ in $\mathcal T$
    \EndIf
\EndFor
\State Store $b_0$ in $\mathcal T$
\Statex
\Statex \textbf{Inference stage}
\Require Image $I$, prompt $x$, frozen model $M$, readout layer $\ell$,
    penalty $\beta$, token table $\mathcal T$
\Ensure Generated sequence $y$
\State During prefill, compute $\dmrr_I(v)$ once
\State Cache $s_I(v)$ for every registered token using
    Eq.~\eqref{eq:suppression-strength}
\While{the response is not complete}
    \State Select candidates $\Cand_t$ from the pre-edit distribution
    \State Apply $z'_t(v)=z_t(v)-\beta s_I(v)$ to
        $v\in\Cand_t\cap\mathcal R$
    \State Decode the next token from the edited logits $\mathbf z'_t$
\EndWhile
\State \Return $y$
\end{algorithmic}
\end{algorithm*}

\subsection{Candidate Construction and Reference Storage}
\label{sec:candidate-construction}

We construct the candidate set from the unedited next-token distribution by sorting tokens by probability and taking $\Cand_t$ as the shortest prefix whose cumulative mass reaches $0.9$.
We expand the prefix to two tokens when the threshold would produce a singleton and truncate it at 50 tokens when the distribution is flat.
The same rule applies during calibration and inference.

Whenever $v\in\Cand_t$ during calibration, the current image-level value $\dmrr_I(v)$ is appended to $D_v$.
The stored table contains $b(v)$ only for registered tokens, together with the shared normalization unit $b_0$ and the registration mask.
Inference needs neither the calibration images nor their hidden states.
We never substitute the pooled value $b_0$ for a missing token reference.

\subsection{Reference-Stability Registration}
\label{sec:registration}

Let $n=|D_v|$ and write the calibration observations for token $v$ as $D_v=\{x_1,\ldots,x_n\}$, with order statistics $x_{(1)}\leq\cdots\leq x_{(n)}$.
We select the largest integer $k$ satisfying the nominal order-statistic rule
\begin{equation}
    1-2\Pr\!\left[Y\leq k-1\right]\geq0.95,
    \qquad Y\sim\operatorname{Bin}(n,\tfrac12),
    \label{eq:median-ci}
\end{equation}
and define the lower and upper endpoints as
\begin{equation}
    \left[b_{\mathrm{lo}}(v),b_{\mathrm{hi}}(v)\right]
    =\left[x_{(k)},x_{(n+1-k)}\right].
    \label{eq:median-ci-interval}
\end{equation}
The resulting endpoints define a nominal 95\% order-statistic reference range.
Calibration observations can repeat within images and decoding trajectories, so we do not interpret this range as a confidence interval with independent sampling or guaranteed 95\% coverage.
We use it only to test whether moving the stored reference between the two order-statistic endpoints materially changes the downstream edit.
If no valid $k$ exists, the token is not registered.
The implied minimum sample size is six, so we do not introduce a separate minimum-count parameter.

We compute the binomial tail exactly for $n\leq200$ and use a continuity-corrected normal approximation for larger collections.
We evaluate the range endpoints in the normalized edit space.
For a candidate reference $b$ and an observed evidence value $x$, define
\begin{equation}
    e(b,x)
    =\clip\!\left(\frac{b-x}{b_0},0,1\right).
\end{equation}
We measure the edit variation induced by the reference range as
\begin{equation}
    \Delta e(v)
    =\frac{1}{|D_v|}\sum_{x\in D_v}
    \left|e\!\left(b_{\mathrm{hi}}(v),x\right)
    -e\!\left(b_{\mathrm{lo}}(v),x\right)\right|.
\end{equation}
We register a token only when $\Delta e(v)<0.5$.
This engineering criterion measures reference stability in the quantity that reaches the decoder rather than attaching a coverage claim to the raw median.
Table~\ref{tab:registration-rules} contrasts this rule with minimum-count registration at the frozen operating point (Appendix~\ref{sec:design-space}).

\subsection{Complete Procedure}

Algorithm~\ref{alg:method} summarizes the complete calibration and inference stages.
Here $\mathcal D_{\mathrm{cal}}$ is the unlabeled calibration image--prompt set, $M$ is the frozen LVLM, and $\mathcal T$ is the frozen token table.
Both stages construct candidate sets with the rule of Appendix~\ref{sec:candidate-construction}, and the reference interval $[b_{\mathrm{lo}}(v),b_{\mathrm{hi}}(v)]$ follows the order-statistic construction of Appendix~\ref{sec:registration}.
At inference, the algorithm computes $\dmrr_I$ once during prefill, caches the suppression strength $s_I(v)$ for registered tokens, and edits only candidates with registered references at each decoding step.

\subsection{Computational Cost}
\label{sec:computational-cost}

\begin{table*}[t]
    \centering
    \small
    \setlength{\tabcolsep}{5pt}
    \begin{tabular}{llr>{\columncolor{gray!15}}rr}
        \toprule
        Group & Quantity & Greedy & \methodname & Overhead \\
        \midrule
        Latency & Cached evidence (mean ratio) & $1.000000\times$ & $1.013323\times$ & $+1.33\%$ \\
                & Online DMRR computation (mean ratio) & $1.000000\times$ & $1.023974\times$ & $+2.40\%$ \\
        \addlinespace
        Memory  & Peak allocated memory (MiB) & 14,353 & 14,397 & $+0.31\%$ \\
        \addlinespace
        Footprint  & Disk, 500 images (MiB) & --- & 183.11 & --- \\
                   & Per image (KiB) & --- & 375.0 & --- \\
        \bottomrule
    \end{tabular}
    \caption{\textbf{Inference-cost accounting.}
    Latency ratios average measurements from three backbones, with 100 prompts
    per backbone. Memory and cache-footprint measurements use \llavaonefive{}.
    Overhead is relative to greedy decoding, and the shaded column is
    \methodname{}.}
    \label{tab:efficiency-accounting}
\end{table*}

For $|P|$ visual positions and a vocabulary of size $|\vocab|$, exact dense ranking requires sorting the vocabulary readout at each position during prefill.
A direct implementation costs $O(|P||\vocab|\log|\vocab|)$ for this one-time operation.
The cached evidence requires $O(|\vocab|)$ memory, and the registered reference table requires $O(|\mathcal R|)$ memory.
Once the candidate set has been formed, the additional per-step editing work is $O(|\Cand_t|)$.
The method introduces no additional backbone forward pass and does not repeat the visual-position projection as the response grows.

Table~\ref{tab:efficiency-accounting} quantifies these costs.
The latency rows average the backbone-specific ratios measured over 100 prompts for each of the three backbones in Table~\ref{tab:chair-amber}.
With the evidence already cached, the mean latency ratio is $\times1.013323$, corresponding to $1.33\%$ overhead over greedy decoding.
When DMRR is computed online during prefill, the mean end-to-end latency ratio is $\times1.023974$, corresponding to $2.40\%$ overhead.
The Footprint rows report the stored evidence cache, which occupies 375~KiB per image across the three cached readout layers.
Deployment reads only the selected layer's slice of this cache.

\section{Additional Quantitative Results}
\label{sec:additional-results}

\subsection{Full CHAIR and AMBER Results}

Tables~\ref{tab:chair-full} and~\ref{tab:amber-full} provide the complete
per-benchmark results, including the metrics omitted from the main-paper
summary table and the \llavanextseven{} backbone.

\begin{table*}[t]
    \centering
    \small
    \setlength{\tabcolsep}{5pt}
    \begin{tabular}{llcccccc>{\columncolor{gray!15}}c}
        \toprule
        Backbone & Metric
        & Base & VCD & OPERA & DoLa & PAI & ReVisiT & \methodname \\
        \midrule
        \llavaonefive & CHAIRs$\downarrow$
        & 52.60 & 59.00 & 51.20 & 53.00 & \textbf{42.40} & 51.40 & \underline{44.80} \\
        & CHAIRi$\downarrow$
        & 15.61 & 17.97 & 14.19 & 15.87 & \underline{13.51} & 15.44 & \textbf{12.67} \\
        & Recall$\uparrow$
        & \textbf{77.25} & 75.26 & 76.63 & 74.95 & 67.06 & \underline{76.94} & 75.26 \\
        & F1$\uparrow$
        & 80.66 & 78.50 & \textbf{80.96} & 79.28 & 75.55 & 80.57 & \underline{80.85} \\
        \midrule
        \qwenvlseven & CHAIRs$\downarrow$
        & 31.60 & 33.80 & \textbf{20.20} & \underline{23.60} & 26.20 & 24.00 & 25.40 \\
        & CHAIRi$\downarrow$
        & 9.58 & 9.65 & \underline{7.01} & 16.76 & 9.61 & \textbf{6.75} & 7.54 \\
        & Recall$\uparrow$
        & \underline{58.17} & 57.49 & 48.54 & 47.30 & 53.20 & 52.27 & \textbf{58.73} \\
        & F1$\uparrow$
        & \underline{70.80} & 70.27 & 63.79 & 60.32 & 66.98 & 66.99 & \textbf{71.83} \\
        \midrule
        \instructblipseven & CHAIRs$\downarrow$
        & 49.60 & 56.20 & 50.80 & 67.00 & 61.00 & \textbf{43.00} & \underline{46.00} \\
        & CHAIRi$\downarrow$
        & \underline{14.03} & 17.67 & 14.77 & 20.26 & 16.70 & 15.02 & \textbf{12.50} \\
        & Recall$\uparrow$
        & 73.09 & 71.41 & \underline{73.65} & \textbf{74.15} & 72.96 & 67.06 & 73.34 \\
        & F1$\uparrow$
        & 79.01 & 76.48 & \underline{79.02} & 76.84 & 77.79 & 74.97 & \textbf{79.80} \\
        \midrule
        \llavanextseven & CHAIRs$\downarrow$
        & 34.00 & 42.40 & \textbf{31.40} & 35.80 & 40.60 & 33.20 & \underline{32.60} \\
        & CHAIRi$\downarrow$
        & 8.67 & 10.87 & \underline{7.98} & 9.54 & 10.62 & 8.75 & \textbf{7.78} \\
        & Recall$\uparrow$
        & 64.82 & 65.94 & 63.46 & \textbf{67.25} & \underline{66.94} & 65.94 & 63.70 \\
        & F1$\uparrow$
        & 75.83 & 75.80 & 75.11 & \textbf{77.14} & 76.55 & \underline{76.56} & 75.36 \\
        \bottomrule
    \end{tabular}
    \caption{\textbf{Full CHAIR results.} Bold and underlined values are
    the best and second-best result within each backbone block, and the
    shaded column is \methodname{}. Arrows indicate the preferred
    direction.}
    \label{tab:chair-full}
\end{table*}

\begin{table*}[t]
    \centering
    \small
    \setlength{\tabcolsep}{3.5pt}
    \begin{tabular}{lcccccccccc}
        \toprule
        & \multicolumn{4}{c}{Gen}
        & \multicolumn{5}{c}{Disc}
        & \multicolumn{1}{c}{Score} \\
        \cmidrule(lr){2-5}\cmidrule(lr){6-10}\cmidrule(lr){11-11}
        Method
        & CHAIR$\downarrow$ & Cover$\uparrow$ & Hal.$\downarrow$ & Cog.$\downarrow$
        & Acc.$\uparrow$ & Prec.$\uparrow$ & Rec.$\uparrow$ & F1$\uparrow$ & Unp.\%$\downarrow$
        & AMBER$\uparrow$ \\
        \midrule
        \llavaonefive
        & \underline{3.69} & 50.47 & 19.02 & 3.82 & 71.96 & \underline{92.65} & 62.73 & 74.81 & \underline{0.08} & 85.56 \\
        \quad  VCD
        & 6.25 & \underline{51.81} & 32.07 & 4.49 & 66.81 & 86.22 & 59.45 & 70.37 & 0.67 & 82.06 \\
        \quad  OPERA
        & 3.70 & 48.79 & \underline{17.03} & \underline{2.70} & \textbf{74.94} & \textbf{92.89} & \underline{67.36} & \textbf{78.09} & \textbf{0.01} & \textbf{87.20} \\
        \quad  DoLa
        & 4.52 & 51.03 & 21.12 & 3.84 & 72.17 & 92.35 & 63.27 & 75.09 & \underline{0.08} & 85.28 \\
        \quad  PAI
        & 4.68 & 43.11 & 19.32 & \textbf{1.97} & 38.29 & 73.46 & 10.86 & 18.93 & 16.66 & 57.12 \\
        \quad  ReVisiT
        & 4.00 & \textbf{55.15} & 24.90 & 2.72 & \underline{73.09} & 85.75 & \textbf{71.25} & \underline{77.80} & 1.75 & \underline{86.90} \\
        \rowcolor{gray!15}
        \quad  \methodname
        & \textbf{2.98} & 50.21 & \textbf{16.43} & 3.13 & 71.96 & \underline{92.65} & 62.73 & 74.81 & \underline{0.08} & 85.91 \\
        \midrule
        \qwenvlseven
        & 4.52 & 55.80 & 20.32 & \underline{1.02} & 77.85 & 84.41 & \underline{91.68} & 87.89 & 9.06 & 91.68 \\
        \quad  VCD
        & 5.43 & \underline{63.30} & 26.99 & 1.77 & 81.77 & 84.18 & 89.29 & 86.66 & 14.78 & 90.61 \\
        \quad  OPERA
        & \textbf{3.51} & \textbf{63.33} & 20.92 & 1.42 & \textbf{83.70} & 85.01 & 91.57 & \textbf{88.17} & \underline{7.94} & \textbf{92.33} \\
        \quad  DoLa
        & 4.47 & 51.71 & 18.92 & 1.56 & 82.55 & \underline{85.17} & 89.21 & 87.15 & 22.00 & 91.33 \\
        \quad  PAI
        & 4.25 & 49.47 & \textbf{18.03} & 1.10 & 81.34 & 84.32 & 88.29 & 86.26 & \textbf{6.89} & 91.00 \\
        \quad  ReVisiT
        & 4.27 & 62.30 & 24.90 & 1.42 & \underline{83.28} & \textbf{85.62} & 89.88 & 87.70 & 17.07 & 91.71 \\
        \rowcolor{gray!15}
        \quad  \methodname
        & \underline{3.80} & 54.65 & \underline{18.82} & \textbf{0.93} & 78.19 & 84.28 & \textbf{92.01} & \underline{87.97} & 8.53 & \underline{92.08} \\
        \midrule
        \instructblipseven
        & 4.64 & \underline{53.74} & 24.40 & 3.80 & \underline{75.91} & 83.49 & \underline{79.43} & \underline{81.40} & 0.08 & \underline{88.38} \\
        \quad  VCD
        & 7.25 & 53.24 & 36.16 & 4.59 & 70.23 & 81.84 & 70.83 & 75.94 & 1.90 & 84.34 \\
        \quad  OPERA
        & 4.81 & 52.08 & 23.80 & 3.53 & 75.90 & 83.66 & 79.10 & 81.32 & \underline{0.04} & 88.25 \\
        \quad  DoLa
        & 7.14 & \textbf{55.26} & 35.86 & 7.09 & 75.76 & 84.10 & 78.23 & 81.06 & 0.07 & 86.96 \\
        \quad  PAI
        & 4.63 & 52.03 & \underline{22.81} & 4.04 & 75.30 & \underline{86.29} & 74.62 & 80.03 & 0.06 & 87.70 \\
        \quad  ReVisiT
        & \underline{4.22} & 50.99 & 22.91 & \textbf{2.13} & 54.36 & \textbf{88.66} & 35.75 & 50.95 & 2.47 & 73.37 \\
        \rowcolor{gray!15}
        \quad  \methodname
        & \textbf{4.05} & 53.28 & \textbf{21.51} & \underline{3.11} & \textbf{75.96} & 83.50 & \textbf{79.48} & \textbf{81.44} & \textbf{0.02} & \textbf{88.69} \\
        \midrule
        \llavanextseven
        & 5.45 & 56.64 & 32.47 & 2.64 & 64.62 & \textbf{99.02} & 56.70 & 72.10 & 19.93 & 83.33 \\
        \quad  VCD
        & 7.55 & 56.21 & 41.33 & 3.11 & 69.46 & 95.28 & 56.75 & 71.13 & \underline{17.33} & 81.79 \\
        \quad  OPERA
        & 5.45 & 54.94 & \underline{31.27} & \underline{2.48} & \textbf{74.93} & 97.91 & \textbf{63.55} & \textbf{77.07} & \textbf{15.39} & \textbf{85.81} \\
        \quad  DoLa
        & 6.19 & \underline{58.24} & 33.27 & 3.03 & \underline{72.21} & 98.68 & \underline{58.88} & \underline{73.76} & 20.05 & 83.79 \\
        \quad  PAI
        & 5.66 & 56.55 & 33.37 & 2.60 & 67.57 & 98.73 & 51.77 & 67.92 & 25.89 & 81.13 \\
        \quad  ReVisiT
        & \underline{4.81} & \textbf{59.53} & 31.47 & 2.58 & 63.03 & 94.61 & 46.92 & 62.73 & 56.25 & 78.96 \\
        \rowcolor{gray!15}
        \quad  \methodname
        & \textbf{4.62} & 56.03 & \textbf{26.49} & \textbf{2.17} & 65.07 & \underline{98.97} & 57.03 & 72.36 & 19.22 & \underline{83.87} \\
        \bottomrule
    \end{tabular}
    \caption{\textbf{Full AMBER results.} Gen and Disc denote the
    generative and discriminative evaluations, respectively. Bold and
    underlined values are the best and second-best result within each
    backbone block, and shaded rows are \methodname{}. Arrows indicate the
    preferred direction.}
    \label{tab:amber-full}
\end{table*}

\subsection{Ablation Definitions and Diagnostics}

\begin{figure*}[t]
    \centering
    \includegraphics[width=\textwidth]{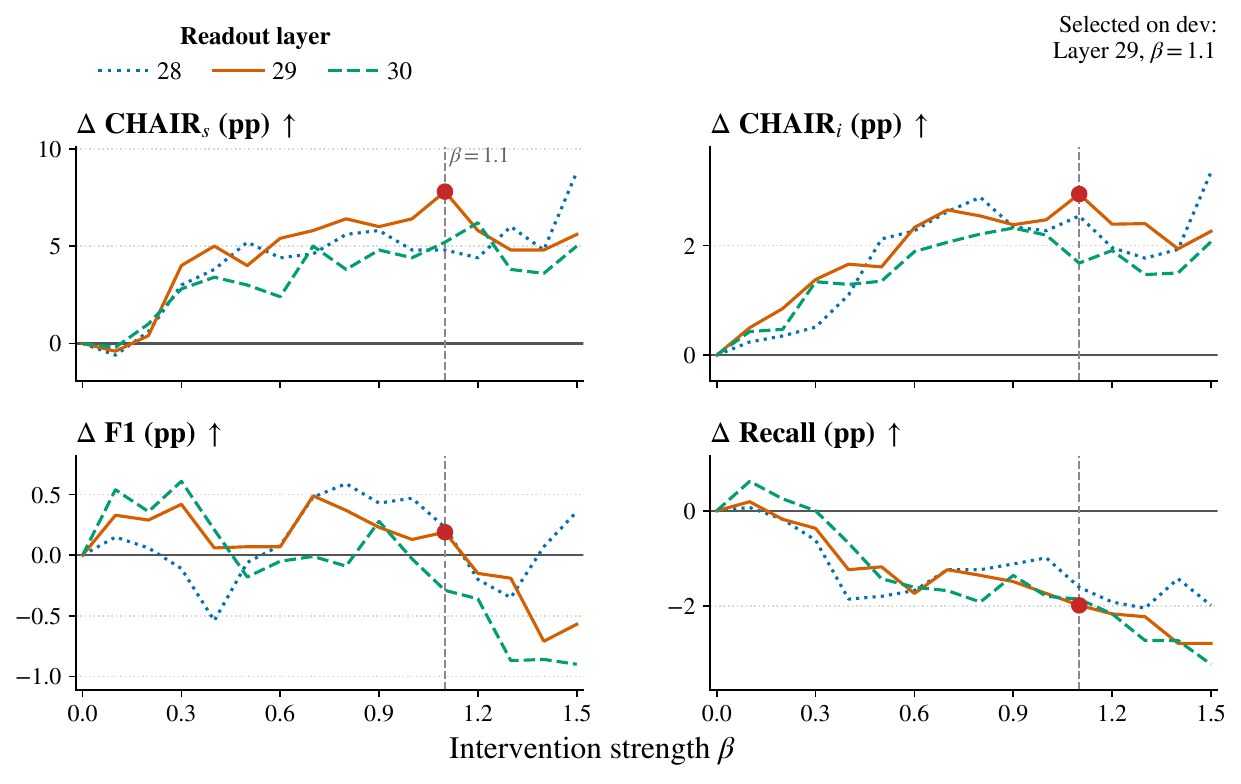}
    \caption{\textbf{Layer-specific sensitivity to the intervention
    strength $\beta$ on the CHAIR validation set.} Circles mark the selected
    operating point ($\ell=29$, $\beta=1.1$).}
    \label{fig:ablation-layer-beta}
\end{figure*}

All ablations use \llavaonefive{} with readout layer $L=29$,
$\beta=1.1$, the same frozen reference-stability table, greedy decoding, and
seed 42.  We evaluate 500 CHAIR validation images and all 1,004 examples in
the AMBER generative split.  Each arm modifies exactly one component of the
full method (Table~\ref{tab:ablation-definitions}).

\begin{table*}[t]
    \centering
    \small
    \begin{tabular}{p{0.18\textwidth}p{0.72\textwidth}}
        \toprule
        Arm & Modification \\
        \midrule
        Global reference & Replace the token-specific reference $b(v)$ with the
        single pooled scale $b_0$. \\
        Shuffled references & Randomly permute $b(v)$ across token identities
        while preserving the multiset of reference values. \\
        Mismatched evidence & Rotate the image-conditioned evidence
        $\dmrr_I(v)$ across examples with a fixed derangement, so no example
        retains its own evidence. \\
        No abstention & Apply the fallback $b_0$ to candidates absent from the
        registered set $\mathcal R$ instead of leaving them unchanged. \\
        Bidirectional update & Remove the suppression-only rectifier
        (the lower clipping bound), permitting evidence to either decrease or
        increase a candidate logit. \\
        Unbounded update & Remove the upper clipping bound on the intervention
        magnitude. \\
        \bottomrule
    \end{tabular}
    \caption{\textbf{Definitions of the component ablations.}}
    \label{tab:ablation-definitions}
\end{table*}

Table~\ref{tab:ablation-chair-detail} reports the CHAIR diagnostics omitted
from the main paper.  The mismatched-evidence arm fires more often than the
full method (87.8\% versus 80.4\%) but performs worse, confirming that the
identity of the visual evidence matters.  The unbounded arm produces 765
bound-escape events and a maximum repetition count of 122.  Its apparent
CHAIR$_S$ gain is therefore inseparable from severe degeneration and coverage
loss.

\begin{table*}[t]
    \centering
    \small
    \setlength{\tabcolsep}{5pt}
    \begin{tabular}{lrrrrrrr}
        \toprule
        Configuration & CHAIR$_S\downarrow$ & CHAIR$_I\downarrow$ & Recall$\uparrow$
        & F1$\uparrow$ & Length & Fire (\%) & Escapes \\
        \midrule
        \rowcolor{gray!15}
        Full method             & 44.8 & 12.67 & 75.3 & 80.85 & 90.3  & 80.4 & 0 \\
        Global reference        & 58.4 & 17.45 & 78.2 & 80.31 & 90.3  & 67.0 & 0 \\
        Shuffled references     & 56.0 & 14.74 & 78.1 & 81.54 & 92.9  & 72.6 & 0 \\
        Mismatched evidence     & 50.2 & 14.45 & 74.6 & 79.73 & 90.2  & 87.8 & 0 \\
        No abstention           & 44.4 & 12.64 & 75.3 & 80.90 & 90.2  & 82.7 & 0 \\
        Bidirectional update    & 50.4 & 16.07 & 75.9 & 79.74 & 92.0  & 98.7 & 0 \\
        Unbounded update        & 39.2 & 29.71 & 67.9 & 69.09 & 106.3 & 79.3 & 765 \\
        \bottomrule
    \end{tabular}
    \caption{\textbf{Detailed CHAIR ablation results.} Fire is the fraction
    of candidate positions at which the intervention is active, and Escapes
    counts updates that would exceed the full method's clipping bound. The
    shaded row is the full method.}
    \label{tab:ablation-chair-detail}
\end{table*}

Table~\ref{tab:image-specificity} expands this counterfactual into a full
image-specificity study at the same frozen operating point.  It replaces the
cached evidence vector while holding every other component fixed.  The
fixed-derangement row is the mismatched-evidence arm above.  A cyclic-shift
swap and an image-blind control ($\dmrr_I\equiv0$) behave the same way, each
recovering less than half of the full method's CHAIR$_S$ reduction and pushing
F1 below base decoding.

\begin{table*}[t]
    \centering
    \small
    \setlength{\tabcolsep}{6pt}
    \begin{tabular}{lrrrr}
        \toprule
        Evidence supplied & CHAIR$_S\downarrow$ & CHAIR$_I\downarrow$
        & Recall$\uparrow$ & F1$\uparrow$ \\
        \midrule
        None (base decoding)                    & 52.6 & 15.61 & 77.3 & 80.66 \\
        \rowcolor{gray!15}
        Own image (\methodname)                 & \textbf{44.8} & \textbf{12.67} & 75.3 & 80.85 \\
        Wrong image, fixed derangement          & 50.2 & 14.45 & 74.6 & 79.73 \\
        Wrong image, cyclic shift               & 50.4 & 15.01 & 75.0 & 79.66 \\
        Image-blind ($\dmrr_I \equiv 0$)        & 49.8 & 13.98 & 74.1 & 79.61 \\
        \bottomrule
    \end{tabular}
    \caption{\textbf{Image-specificity study on \llavaonefive{}.} Each row
    replaces the cached evidence vector with the stated substitute. The
    fixed-derangement row is the mismatched-evidence arm of
    Table~\ref{tab:signal-ablation}. Arrows indicate the preferred direction,
    and the shaded row is the unmodified method.}
    \label{tab:image-specificity}
\end{table*}

On AMBER generation, the well-behaved arms show the same overall degradation
pattern as on CHAIR (Table~\ref{tab:ablation-amber-detail}).
Abstention is a conservative safeguard rather than a performance component.
It withholds the edit for tokens whose reference estimate is unstable, so one
unreliable reference does not distort otherwise sound generation.  The arm
that removes it touches only about 3.5\% of candidate positions and shifts
CHAIR$_S$ by $0.4$ points and CHAIR F1 by $0.05$ points.  On the
discriminative split, removing abstention raises F1 from 74.81 to 76.20 while
moving the predicted yes/no counts from 7,822/6,383 to 7,420/6,769.  The
prediction shift favors ``no'', the majority label, and lowers precision as
recall rises.  We therefore read this F1 movement as majority-class drift
rather than improved visual grounding, and such marginal movement does not
outweigh the safeguard.

\begin{table*}[t]
    \centering
    \small
    \setlength{\tabcolsep}{3.5pt}
    \begin{tabular}{lcccccccccc}
        \toprule
        & \multicolumn{4}{c}{Gen}
        & \multicolumn{5}{c}{Disc}
        & \multicolumn{1}{c}{Score} \\
        \cmidrule(lr){2-5}\cmidrule(lr){6-10}\cmidrule(lr){11-11}
        Configuration
        & CHAIR$\downarrow$ & Cover$\uparrow$ & Hal.$\downarrow$ & Cog.$\downarrow$
        & Acc.$\uparrow$ & Prec.$\uparrow$ & Rec.$\uparrow$ & F1$\uparrow$ & Unp.\%$\downarrow$
        & AMBER$\uparrow$ \\
        \midrule
        \rowcolor{gray!15}
        Full method
        & 2.98 & 50.21 & 16.43 & 3.13 & 71.96 & 92.65 & 62.73 & 74.81 & 0.08 & 85.91 \\
        Global reference
        & 4.10 & 52.78 & 21.91 & 3.51 & 71.98 & 92.65 & 62.75 & 74.82 & 0.05 & 85.36 \\
        Shuffled references
        & 4.11 & 51.63 & 21.51 & 4.14 & 71.98 & 92.65 & 62.75 & 74.82 & 0.05 & 85.36 \\
        Mismatched evidence
        & 3.70 & 49.35 & 18.03 & 3.15 & 71.96 & 92.65 & 62.73 & 74.81 & 0.08 & 85.56 \\
        No abstention
        & 2.97 & 50.37 & 16.14 & 3.01 & 72.82 & 91.17 & 65.46 & 76.20 & 0.19 & 86.62 \\
        Bidirectional update
        & 3.72 & 51.22 & 19.02 & 2.94 & 71.95 & 92.65 & 62.72 & 74.80 & 0.08 & 85.54 \\
        Unbounded update
        & 3.44 & 48.26 & 16.14 & 2.62 & 71.96 & 92.65 & 62.73 & 74.81 & 0.08 & 85.69 \\
        \bottomrule
    \end{tabular}
    \caption{\textbf{Detailed AMBER ablation results.} Gen and Disc denote
    the generative and discriminative evaluations, respectively. The shaded
    row is the full method. Arrows indicate the preferred direction.}
    \label{tab:ablation-amber-detail}
\end{table*}

Table~\ref{tab:layer-beta-grid} lists the full layer~$\times$~$\beta$ grid
behind Figure~\ref{fig:ablation-layer-beta}, covering layers 28--30 and
$\beta$ from 0 to 1.5.  We selected the frozen
operating point ($\ell=29$, $\beta=1.1$) on the seed-42 COCO train2014
tuning subset before these validation runs. It is not the per-cell optimum on
this split.

\begin{table*}[t]
    \centering
    \small
    \setlength{\tabcolsep}{4pt}
    \begin{tabular}{r rrr rrr rrr}
        \toprule
        & \multicolumn{3}{c}{$\ell=28$} & \multicolumn{3}{c}{$\ell=29$}
        & \multicolumn{3}{c}{$\ell=30$} \\
        \cmidrule(lr){2-4}\cmidrule(lr){5-7}\cmidrule(lr){8-10}
        $\beta$ & $C_S\downarrow$ & $C_I\downarrow$ & F1$\uparrow$
        & $C_S\downarrow$ & $C_I\downarrow$ & F1$\uparrow$
        & $C_S\downarrow$ & $C_I\downarrow$ & F1$\uparrow$ \\
        \midrule
        0.0 & 52.6 & 15.61 & 80.66 & 52.6 & 15.61 & 80.66 & 52.6 & 15.61 & 80.66 \\
        0.1 & 53.2 & 15.37 & 80.81 & 53.0 & 15.11 & 80.99 & 52.8 & 15.18 & 81.20 \\
        0.2 & 52.0 & 15.26 & 80.72 & 52.2 & 14.76 & 80.95 & 51.6 & 15.14 & 81.02 \\
        0.3 & 49.6 & 15.10 & 80.55 & 48.6 & 14.23 & 81.08 & 49.8 & 14.27 & 81.27 \\
        0.4 & 48.8 & 14.51 & 80.12 & 47.6 & 13.95 & 80.72 & 49.2 & 14.32 & 80.87 \\
        0.5 & 47.4 & 13.49 & 80.60 & 48.6 & 14.00 & 80.73 & 49.6 & 14.26 & 80.48 \\
        0.6 & 48.2 & 13.34 & 80.74 & 47.2 & 13.28 & 80.73 & 50.2 & 13.72 & 80.61 \\
        0.7 & 48.0 & 12.99 & 81.14 & 46.8 & 12.96 & 81.15 & 47.6 & 13.55 & 80.65 \\
        0.8 & 47.0 & 12.73 & 81.25 & 46.2 & 13.07 & 81.03 & 48.8 & 13.40 & 80.57 \\
        0.9 & 46.8 & 13.27 & 81.09 & 46.6 & 13.23 & 80.89 & 47.8 & 13.29 & 80.94 \\
        1.0 & 47.8 & 13.34 & 81.13 & 46.2 & 13.14 & 80.79 & 48.2 & 13.42 & 80.63 \\
        1.1 & 47.8 & 13.07 & 80.89 & \cellcolor{gray!15}44.8 & \cellcolor{gray!15}12.67 & \cellcolor{gray!15}80.85 & 47.4 & 13.93 & 80.37 \\
        1.2 & 48.2 & 13.65 & 80.46 & 46.8 & 13.22 & 80.51 & 46.4 & 13.70 & 80.30 \\
        1.3 & 46.6 & 13.84 & 80.31 & 47.8 & 13.21 & 80.47 & 48.8 & 14.14 & 79.79 \\
        1.4 & 47.8 & 13.68 & 80.73 & 47.8 & 13.67 & 79.95 & 49.0 & 14.11 & 79.80 \\
        1.5 & 43.8 & 12.27 & 81.02 & 47.0 & 13.35 & 80.09 & 47.6 & 13.54 & 79.76 \\
        \bottomrule
    \end{tabular}
    \caption{\textbf{Full layer~$\times$~$\beta$ grid on the CHAIR
    validation split (\llavaonefive{}).} $C_S$ and $C_I$ denote CHAIR$_S$
    and CHAIR$_I$, the $\beta{=}0$ rows equal base decoding, and the shaded
    cell is the frozen operating point.}
    \label{tab:layer-beta-grid}
\end{table*}

\subsection{Design-Space and Data Robustness}
\label{sec:design-space}

Table~\ref{tab:registration-rules} contrasts the reference-stability rule of
Section~\ref{sec:calib} with minimum-count registration ($n(v)\geq5$ or
$n(v)\geq10$) under an otherwise frozen configuration.  Every rule recovers
most of the CHAIR improvement over base decoding, and the three rules stay
within $1.2$ CHAIR$_S$ points of one another on every backbone.  The
reference-stability rule also avoids the AMBER F1 drop that the $n(v)\geq5$
rule shows on \instructblipseven{}.  A fixed occurrence
cutoff, however, is an external constant.  The counts it gates on grow with
the calibration budget, and the evidence scale behind each reference varies
across backbones with different visual-token counts, so the same threshold
acts differently as either factor changes.  We use the reference-stability
rule because it gates each token on the normalized edit that reaches the
decoder, which adapts to both factors without a separate minimum-count
parameter (Appendix~\ref{sec:registration}).

\begin{table*}[t]
    \centering
    \small
    \setlength{\tabcolsep}{4pt}
    \begin{tabular}{llrrrrrr}
        \toprule
        Backbone & Registration rule & $|\mathcal R|$
        & CHAIR$_S\downarrow$ & CHAIR$_I\downarrow$ & Recall$\uparrow$ & F1$\uparrow$
        & AMBER-disc F1$\uparrow$ \\
        \midrule
        \llavaonefive & none (base)          & --    & 52.6 & 15.61 & 77.3 & 80.66 & 74.81 \\
        & count $n(v)\geq5$                  & 3,252 & 44.6 & 12.69 & 75.4 & 80.91 & 74.81 \\
        & count $n(v)\geq10$                 & 2,476 & 44.8 & 12.70 & 75.4 & 80.91 & 74.81 \\
        \rowcolor{gray!15}
        & reference stability ($\Delta e<0.5$) & 2,580 & 44.8 & 12.67 & 75.3 & 80.85 & 74.81 \\
        \midrule
        \instructblipseven & none (base)     & --    & 49.6 & 14.03 & 73.1 & 79.01 & 81.40 \\
        & count $n(v)\geq5$                  & 3,524 & 46.2 & 12.50 & 73.3 & 79.80 & 78.51 \\
        & count $n(v)\geq10$                 & 2,714 & 46.2 & 12.54 & 73.3 & 79.74 & 81.44 \\
        \rowcolor{gray!15}
        & reference stability ($\Delta e<0.5$) & 2,708 & 46.0 & 12.50 & 73.3 & 79.80 & 81.44 \\
        \midrule
        \llavanextseven & none (base)        & --    & 34.0 &  8.67 & 64.8 & 75.83 & 72.10 \\
        & count $n(v)\geq5$                  & 4,791 & 32.2 &  7.66 & 64.3 & 75.78 & 72.36 \\
        & count $n(v)\geq10$                 & 3,572 & 32.2 &  7.72 & 64.1 & 75.68 & 72.36 \\
        \rowcolor{gray!15}
        & reference stability ($\Delta e<0.5$) & 3,734 & 32.6 &  7.78 & 63.7 & 75.36 & 72.36 \\
        \midrule
        \qwenvlseven & none (base)           & --    & 31.6 &  9.58 & 58.2 & 70.80 & 87.89 \\
        & count $n(v)\geq5$                  & 9,286 & 26.2 &  7.64 & 58.4 & 71.57 & 87.97 \\
        & count $n(v)\geq10$                 & 6,458 & 26.6 &  7.87 & 58.4 & 71.50 & 87.97 \\
        \rowcolor{gray!15}
        & reference stability ($\Delta e<0.5$) & 7,028 & 25.4 &  7.54 & 58.7 & 71.83 & 87.97 \\
        \bottomrule
    \end{tabular}
    \caption{\textbf{Design-space contrast of registration rules.}
    $|\mathcal R|$ is the number of registered tokens. Arrows indicate the
    preferred direction, and the shaded rows are the reference-stability rule
    used by \methodname{}.}
    \label{tab:registration-rules}
\end{table*}

Table~\ref{tab:calibration-size} lists the calibration-size sweep behind
Figure~\ref{fig:calibration-size}, including recall.  The nested subsets
share the frozen operating point, and the 500-image row reuses the deployed
table and evaluation output (Appendix~\ref{sec:implementation-details}).

\begin{table*}[t]
    \centering
    \small
    \setlength{\tabcolsep}{6pt}
    \begin{tabular}{rrrrr}
        \toprule
        Calibration images & CHAIR$_S\downarrow$ & CHAIR$_I\downarrow$
        & Recall$\uparrow$ & F1$\uparrow$ \\
        \midrule
        None (base decoding) & 52.6 & 15.61 & 77.3 & 80.66 \\
        10    & 48.6 & 14.29 & 74.64 & 79.80 \\
        25    & 48.0 & 14.04 & 75.14 & 80.19 \\
        50    & 48.2 & 13.99 & 76.01 & 80.70 \\
        100   & 44.8 & 13.29 & 75.76 & 80.87 \\
        200   & 45.2 & 12.82 & 75.14 & 80.71 \\
        \rowcolor{gray!15}
        500   & 44.8 & 12.67 & 75.26 & 80.85 \\
        700   & 44.4 & 12.62 & 75.82 & 81.19 \\
        1,000 & 43.8 & 12.35 & 76.07 & 81.45 \\
        \bottomrule
    \end{tabular}
    \caption{\textbf{Calibration-size sweep on \llavaonefive{}.} Arrows
    indicate the preferred direction, and the shaded row is the deployed
    500-image configuration.}
    \label{tab:calibration-size}
\end{table*}

Table~\ref{tab:calibration-corpus} refits the calibration table on different
unlabeled corpora at the frozen operating point.  GQA calibration \citep{hudson2019gqa} matches the
MS COCO default on CHAIR$_S$, and AMBER and pooled calibration remain well
below base decoding.  The method therefore does not require calibrating on
the evaluation distribution, although corpus shift can move the operating
point.

\begin{table*}[t]
    \centering
    \small
    \setlength{\tabcolsep}{5pt}
    \begin{tabular}{lrrrrrr}
        \toprule
        Calibration corpus & Images & $|\mathcal R|$
        & CHAIR$_S\downarrow$ & CHAIR$_I\downarrow$ & Recall$\uparrow$ & F1$\uparrow$ \\
        \midrule
        None (base decoding)     & --    & --    & 52.6 & 15.61 & 77.3 & 80.66 \\
        \rowcolor{gray!15}
        MS COCO (default)        & 500   & 2,580 & 44.8 & 12.67 & 75.3 & 80.85 \\
        GQA                      & 500   & 2,429 & 44.6 & 13.15 & 75.4 & 80.71 \\
        AMBER                    & 1,004 & 2,982 & 48.0 & 12.97 & 75.6 & 80.93 \\
        Pooled (COCO+GQA+AMBER)  & 2,004 & 3,900 & 46.6 & 13.22 & 75.9 & 81.00 \\
        \bottomrule
    \end{tabular}
    \caption{\textbf{Calibration-corpus transfer on \llavaonefive{}.}
    $|\mathcal R|$ is the number of registered tokens. Arrows indicate the
    preferred direction, and the shaded row is the default configuration.}
    \label{tab:calibration-corpus}
\end{table*}

\subsection{Scaling across Model Sizes and Architectures}
\label{sec:model-scaling}

We further evaluate \methodname{} across 11 models spanning six architecture
families and model sizes from 7B to 32B. Results use the 500 examples in CHAIR
val and reference-stability registration with the fixed $b_0$ scale. CHAIR$_S$,
CHAIR$_I$, recall, and F1 are reported on a 0--100 scale.

The scaling study assigns operating points by family. A 13B or 32B model
reuses the $\beta$ of its 7B backbone in Table~\ref{tab:operating-points}
and places its readout layer at the same depth from the final decoder
layer. We tune the two Gemma-3 models separately using a restricted search.
For each model, we fix the readout to the third-to-last decoder layer and
compare only $\beta\in\{0.5,1.0\}$ on the seed-42 COCO train2014 tuning
subset. We select the setting by jointly reviewing CHAIR$_S$, CHAIR$_I$,
recall, and F1. This yields $(\ell,\beta)=(45,1.0)$ for Gemma-3-12B and
$(\ell,\beta)=(59,0.5)$ for Gemma-3-27B. For
InternVL2.5-8B and InternVL2.5-26B, we use the same operating-point search
described in Appendix~\ref{sec:implementation-details}. On the seed-42
COCO train2014 tuning subset, we pair the final three decoder layers with
$\beta\in\{0.5,0.7,0.9,1.1,1.3\}$ and then evaluate one additional midpoint
at all three layers. We jointly review CHAIR$_S$, CHAIR$_I$, recall, and F1
over the resulting 18 configurations. This procedure yields
$(\ell,\beta)=(31,1.1)$ for the 8B model and $(\ell,\beta)=(45,1.1)$ for
the 26B model. We freeze these settings before evaluating the models on the
separate 500-image COCO val2014 subset reported in
Table~\ref{tab:model-scaling}.

As shown in Table~\ref{tab:model-scaling}, the selected \methodname{} settings
reduce both CHAIR$_S$ and CHAIR$_I$ for every model.  This trend holds across
all tested sizes and architectures.  F1 improves for six of the 11
models.  For the remaining models, F1 stays within $0.7$ points of base
decoding, and each decrease occurs where recall also falls.

\begin{table*}[t]
    \centering
    \begin{tabular}{llrrrr}
        \toprule
        Model & Method & CHAIR$_S\downarrow$
        & CHAIR$_I\downarrow$ & Recall$\uparrow$ & F1$\uparrow$ \\
        \midrule
        InstructBLIP-7B  & Greedy        & 49.60 & 14.03 & 73.09 & 79.01 \\
                          & \methodname{} & \cellcolor{gray!15}46.00 & \cellcolor{gray!15}12.50 & \cellcolor{gray!15}73.30 & \cellcolor{gray!15}79.80 \\
        LLaVA-1.5-7B     & Greedy        & 52.60 & 15.61 & 77.25 & 80.66 \\
                          & \methodname{} & \cellcolor{gray!15}44.80 & \cellcolor{gray!15}12.67 & \cellcolor{gray!15}75.30 & \cellcolor{gray!15}80.85 \\
        LLaVA-NeXT-7B    & Greedy        & 34.00 &  8.67 & 64.82 & 75.83 \\
                          & \methodname{} & \cellcolor{gray!15}32.60 & \cellcolor{gray!15} 7.78 & \cellcolor{gray!15}63.70 & \cellcolor{gray!15}75.36 \\
        Qwen2.5-VL-7B    & Greedy        & 31.60 &  9.58 & 58.17 & 70.80 \\
                          & \methodname{} & \cellcolor{gray!15}25.40 & \cellcolor{gray!15} 7.54 & \cellcolor{gray!15}58.70 & \cellcolor{gray!15}71.83 \\
        InternVL2.5-8B   & Greedy        & 32.40 &  8.21 & 69.36 & 79.01 \\
                          & \methodname{} & \cellcolor{gray!15}30.00 & \cellcolor{gray!15} 7.51 & \cellcolor{gray!15}68.18 & \cellcolor{gray!15}78.50 \\
        Gemma-3-12B      & Greedy        & 49.80 &  8.95 & 70.60 & 79.53 \\
                          & \methodname{} & \cellcolor{gray!15}46.80 & \cellcolor{gray!15} 8.53 & \cellcolor{gray!15}70.04 & \cellcolor{gray!15}79.33 \\
        InstructBLIP-13B & Greedy        & 58.00 & 17.23 & 70.79 & 76.31 \\
                          & \methodname{} & \cellcolor{gray!15}54.60 & \cellcolor{gray!15}15.95 & \cellcolor{gray!15}70.04 & \cellcolor{gray!15}76.41 \\
        LLaVA-1.5-13B    & Greedy        & 51.80 & 14.76 & 77.50 & 81.19 \\
                          & \methodname{} & \cellcolor{gray!15}45.00 & \cellcolor{gray!15}11.70 & \cellcolor{gray!15}75.95 & \cellcolor{gray!15}81.66 \\
        InternVL2.5-26B  & Greedy        & 31.20 &  8.16 & 69.98 & 79.43 \\
                          & \methodname{} & \cellcolor{gray!15}28.60 & \cellcolor{gray!15} 7.19 & \cellcolor{gray!15}69.48 & \cellcolor{gray!15}79.47 \\
        Gemma-3-27B      & Greedy        & 53.00 &  9.51 & 72.03 & 80.21 \\
                          & \methodname{} & \cellcolor{gray!15}50.60 & \cellcolor{gray!15} 9.17 & \cellcolor{gray!15}70.79 & \cellcolor{gray!15}79.57 \\
        Qwen2.5-VL-32B   & Greedy        & 59.40 & 10.93 & 75.33 & 81.62 \\
                          & \methodname{} & \cellcolor{gray!15}54.60 & \cellcolor{gray!15}10.15 & \cellcolor{gray!15}74.58 & \cellcolor{gray!15}81.51 \\
        \bottomrule
    \end{tabular}
    \caption{\textbf{CHAIR results across model sizes and architecture
    families.} Shaded cells mark \methodname{}.}
    \label{tab:model-scaling}
\end{table*}

\section{Qualitative Results and Error Analysis}
\label{sec:additional-examples}

This appendix complements the quantitative results with recorded case
studies. It unfolds the intervention arithmetic at a single decoding step,
presents additional side-by-side examples from CHAIR and AMBER, and
illustrates failure modes that appear when individual design components are
removed. Throughout, \textcolor{red!70!black}{\textbf{red}} marks object
mentions that the benchmark scorer counts as hallucinated, and
\textcolor{blue!70!black}{\textbf{blue}} marks objects that the scorer
matches to the reference annotations.
The rank--probability output comparison appears separately in
Appendix~\ref{sec:detection-not-mitigation}.

\newcommand{\halluc}[1]{\textcolor{red!70!black}{\textbf{#1}}}
\newcommand{\grounded}[1]{\textcolor{blue!70!black}{\textbf{#1}}}
\newcommand{\exampleheader}[2]{%
    \noindent\colorbox{black!7}{%
        \parbox{\dimexpr\linewidth-2\fboxsep\relax}{%
            \textbf{#1\enspace #2}}}}
\newcommand{\answerbox}[3]{%
    \noindent\colorbox{#1}{%
        \parbox{\dimexpr\linewidth-2\fboxsep\relax}{%
            \textbf{#2}\quad #3}}}

% Keep full-width qualitative figures top-aligned on float-only appendix pages.
\makeatletter
\setlength{\@dblfptop}{0pt}
\setlength{\@dblfpbot}{0pt plus 1fil}
\makeatother

\subsection{Mechanism Case Study}
\label{sec:mechanism-case}

\begin{figure*}[t]
    \centering
    \exampleheader{(a)}{\llavaonefive}
    \vspace{4pt}

    \begin{minipage}[c]{0.22\textwidth}
        \centering
        \includegraphics[width=\linewidth]{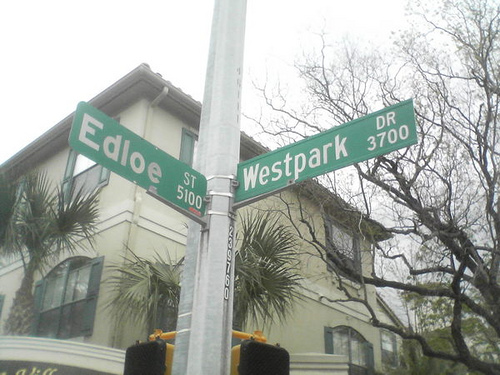}
    \end{minipage}\hfill
    \begin{minipage}[c]{0.75\textwidth}
        \centering
        \includegraphics[width=\linewidth]{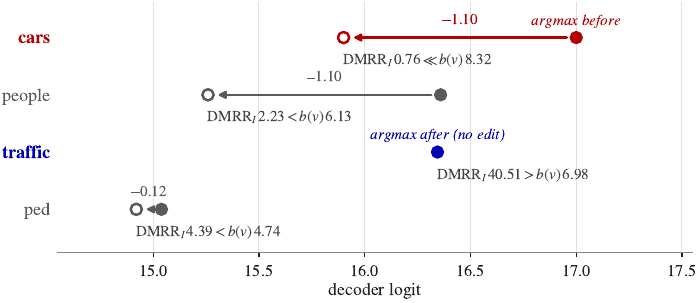}
    \end{minipage}

    \vspace{9pt}
    \exampleheader{(b)}{\llavanextseven}
    \vspace{4pt}

    \begin{minipage}[c]{0.22\textwidth}
        \centering
        \includegraphics[width=\linewidth]{figure/raw_img/COCO_val2014_000000000428.jpg}
    \end{minipage}\hfill
    \begin{minipage}[c]{0.75\textwidth}
        \centering
        \includegraphics[width=\linewidth]{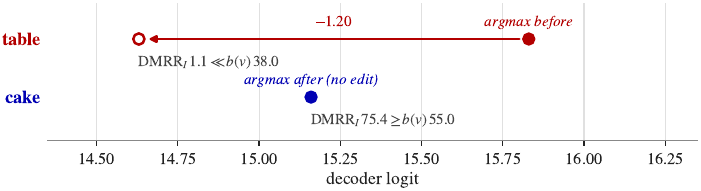}
    \end{minipage}
    \caption{\textbf{Mechanism case studies read from recorded traces.} Each
    panel plots decoder logits before (filled) and after (open) the edit at
    one recorded decoding step, and each arrow spans the bounded suppression
    $\beta s_I(v)$. The annotation under each candidate compares its evidence
    $\dmrr_I(v)$ with its reference $b(v)$ in units of $10^{-3}$. Panel (a)
    shows \llavaonefive{} at step 71 with four candidates from
    Table~\ref{tab:mechanism-case} that span full, proportional, and zero
    suppression. Panel (b) shows \llavanextseven{} ($\beta{=}1.2$, layer-30
    readout) at step 60. Red marks the hallucinated pre-edit argmax, and blue
    marks the grounded token that becomes the post-edit argmax.}
    \label{fig:mechanism-case}
\end{figure*}

Table~\ref{tab:mechanism-case} and Figure~\ref{fig:mechanism-case} unfold the
intervention arithmetic at a single recorded decoding step of
\llavaonefive{}. At this step the base model's argmax is the object token
\texttt{cars}, which is absent from the image. Its evidence readout falls far
below the token-specific reference ($\dmrr_I(v) \ll b(v)$), the deficit
saturates the normalization ($s_I(v){=}1$), and the token is displaced by the
full bound $\beta$. The evidence-supported candidate \texttt{traffic} sits
above its own reference, receives no edit, and becomes the post-edit argmax,
so the caption continues with the traffic lights that the image does contain.
The same table shows the two other regimes at work within a single candidate
set: registered tokens whose evidence clears their reference are left
untouched, and a partially supported candidate receives a proportional
intermediate penalty. Figure~\ref{fig:mechanism-case}(b) shows a
complementary case for \llavanextseven{} in which the suppressed hallucinated
object is directly replaced by a grounded one. All values come from recorded
traces of the deployed configuration, and no quantity is recomputed post hoc.

\begin{table*}[t]
    \centering
    \small
    \setlength{\tabcolsep}{6pt}
    \begin{tabular}{lrrrrrrr}
        \toprule
        & & \multicolumn{3}{c}{Evidence ($\times 10^{-3}$)} & & & \\
        \cmidrule(lr){3-5}
        Candidate & $z_t(v)$ & $\dmrr_I(v)$ & $b(v)$ & Deficit & $s_I(v)$ & $\beta\,s_I(v)$ & $z'_t(v)$ \\
        \midrule
        \textbf{\texttt{cars}} & 17.000 & 0.758 & 8.321 & 7.563 & 1.000 & 1.100 & 15.900 \\
        \texttt{people} & 16.359 & 2.225 & 6.128 & 3.903 & 1.000 & 1.100 & 15.259 \\
        \rowcolor{gray!15}
        \texttt{traffic} & 16.344 & 40.513 & 6.984 & 0.000 & 0.000 & 0.000 & 16.344 \\
        \texttt{p} & 15.891 & 7.390 & 23.583 & 16.192 & 1.000 & 1.100 & 14.791 \\
        \texttt{pal} & 15.703 & 72.867 & 124.762 & 51.895 & 1.000 & 1.100 & 14.603 \\
        \texttt{stop} & 15.063 & 10.986 & 5.882 & 0.000 & 0.000 & 0.000 & 15.063 \\
        \texttt{ped} & 15.039 & 4.393 & 4.743 & 0.350 & 0.108 & 0.119 & 14.920 \\
        \texttt{clock} & 14.953 & 0.760 & 7.693 & 6.933 & 1.000 & 1.100 & 13.853 \\
        \bottomrule
    \end{tabular}
    \caption{\textbf{Per-candidate arithmetic of \methodname{} at a single
    recorded decoding step.} The step uses the deployed \llavaonefive{}
    configuration of Table~\ref{tab:operating-points}. The table lists the
    eight candidates with the highest pre-edit logits among the $K{=}22$
    candidates, including the pre-edit argmax (bold) and the emitted token
    (highlighted). The columns reproduce the arithmetic of
    Eqs.~\eqref{eq:normalized-edit}, \eqref{eq:suppression-strength},
    and~\eqref{eq:shift} from the recorded trace.}
    \label{tab:mechanism-case}
\end{table*}

\subsection{Additional CHAIR Examples}
\label{sec:chair-examples}

\begin{figure*}[t]
    \centering
    \exampleheader{(a)}{\instructblipseven}
    \vspace{5pt}

    \begin{minipage}[t]{0.24\textwidth}
        \vspace{0pt}
        \centering
        \includegraphics[width=\linewidth]{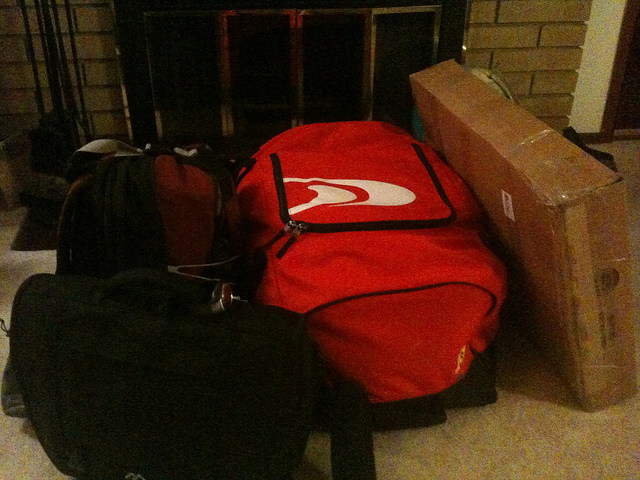}
    \end{minipage}\hfill
    \begin{minipage}[t]{0.73\textwidth}
        \vspace{0pt}
        \small
        \answerbox{red!3}{Base}{The image features a living room with a fireplace, where
        a red duffel bag is placed on the floor next to a \halluc{chair}. There
        are several other bags and suitcases scattered around the room,
        including a black bag and a brown bag. The room also contains a
        \halluc{television}, which is placed on a \halluc{table} in the center of
        the room. A \halluc{potted plant} can be seen on the left side of the
        room, and a \halluc{book} is placed on the right side of the room. The
        overall scene suggests that the room is being prepared for a trip or a
        move.}

        \vspace{5pt}
        \answerbox{blue!3}{$+$\methodname}{The image features a living room with a
        fireplace, where several bags and suitcases are placed on the floor next
        to the fireplace. There are three bags visible, with one of them being a
        red one, and the other two being black. The bags are placed close to each
        other, with one bag located closer to the fireplace and the other two
        bags further away. There is also a \grounded{box} placed on the floor near the
        fireplace, which is likely to contain additional items. The room appears
        to be well-organized, with the bags and box neatly arranged.}
    \end{minipage}

    \vspace{10pt}
    \exampleheader{(b)}{\llavaonefive}
    \vspace{5pt}

    \begin{minipage}[t]{0.24\textwidth}
        \vspace{0pt}
        \centering
        \includegraphics[width=\linewidth]{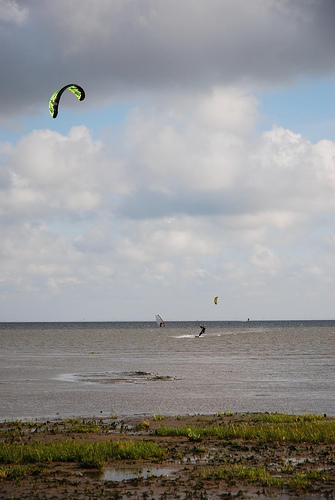}
    \end{minipage}\hfill
    \begin{minipage}[t]{0.73\textwidth}
        \vspace{0pt}
        \small
        \answerbox{red!3}{Base}{The image captures a lively scene of people enjoying
        water sports on a large body of water. There are two people in the
        water, one of them is parasailing, while the other is windsurfing. The
        parasailer is being pulled by a \halluc{boat}, and the windsurfer is
        skillfully navigating the water. In the sky above, a kite can be seen
        flying, adding to the excitement of the scene. The water is filled with
        various \halluc{boats}, including a couple of smaller \halluc{boats}
        near the shore and a larger \halluc{boat} further out in the water. The
        presence of multiple \halluc{boats} and people participating in water
        sports creates a vibrant and active atmosphere.}

        \vspace{5pt}
        \answerbox{blue!3}{$+$\methodname}{The image captures a beautiful scene of a
        person kiteboarding in the ocean, surrounded by a few other
        \grounded{kiteboarders}. The person in the center of the scene is skillfully riding
        a \grounded{kiteboard}, while others are also enjoying the water nearby. In the
        sky, there are two \grounded{kites} visible, one towards the left side and the other
        towards the right. The \grounded{kites} are flying high above the water, adding to
        the excitement of the kiteboarding activity. The overall atmosphere of
        the scene is lively and full of adventure.}
    \end{minipage}
    \vspace{10pt}
    \exampleheader{(c)}{\instructblipseven}
    \vspace{5pt}

    \begin{minipage}[t]{0.24\textwidth}
        \vspace{0pt}
        \centering
        \includegraphics[width=\linewidth]{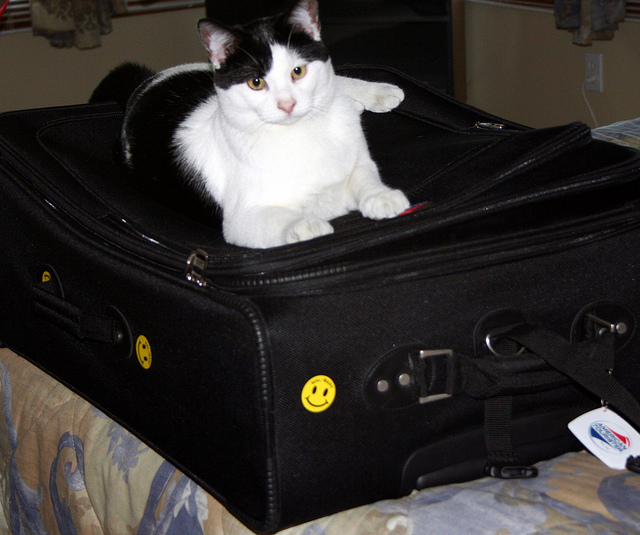}
    \end{minipage}\hfill
    \begin{minipage}[t]{0.73\textwidth}
        \vspace{0pt}
        \small
        \answerbox{red!3}{Base}{The image features a black and white cat sitting on top
        of a black suitcase placed on a bed. The cat is lying on the suitcase,
        which is open and partially visible. The cat's body is stretched out,
        and its head is resting on the top of the suitcase. The suitcase is
        placed on the bed, which is covered with a blanket. There are several
        other items on the bed, including a \halluc{book}, a \halluc{cup}, and a
        \halluc{bottle}. The cat seems to be enjoying its time on the suitcase,
        possibly taking a nap or simply relaxing.}

        \vspace{5pt}
        \answerbox{blue!3}{$+$\methodname}{The image features a black and white cat
        sitting on top of a black suitcase placed on a bed. The cat is lying on
        the suitcase, which is open and partially visible on the bed. The cat is
        positioned in the center of the suitcase, with its head and body
        stretched out on top of it. The suitcase is placed on a bed, which is
        covered with a blanket and has a \grounded{pillow} on the left side. The cat seems
        to be enjoying its resting spot on the suitcase, possibly waiting for
        its owner to return.}
    \end{minipage}
    \caption{\textbf{Qualitative CHAIR examples of three common
    hallucination patterns.} Panels (a), (b), and (c) show contextual scene
    completion, repeated object generation, and peripheral errors,
    respectively. Red marks CHAIR-hallucinated objects, and blue marks
    ground-truth objects.}
    \label{fig:chair-examples-1}
\end{figure*}

Figure~\ref{fig:chair-examples-1} provides additional examples from CHAIR.
We compare the response from the base model with the response obtained using
\methodname{} under the same image and prompt. These examples cover different
backbones and common hallucination patterns, including contextual scene
completion, repeated object generation, and removal of peripheral errors
while preserving recognized objects.

Across these examples, \methodname{} selectively suppresses the flagged
object mentions while retaining or introducing descriptions of matched
objects. This behavior applies to both isolated object mentions and
relations built around those objects.

\subsection{AMBER Examples}
\label{sec:amber-examples}

\begin{figure*}[t]
    \centering
    \exampleheader{(a)}{\llavaonefive}
    \vspace{5pt}

    \begin{minipage}[t]{0.24\textwidth}
        \vspace{0pt}
        \centering
        \includegraphics[width=\linewidth]{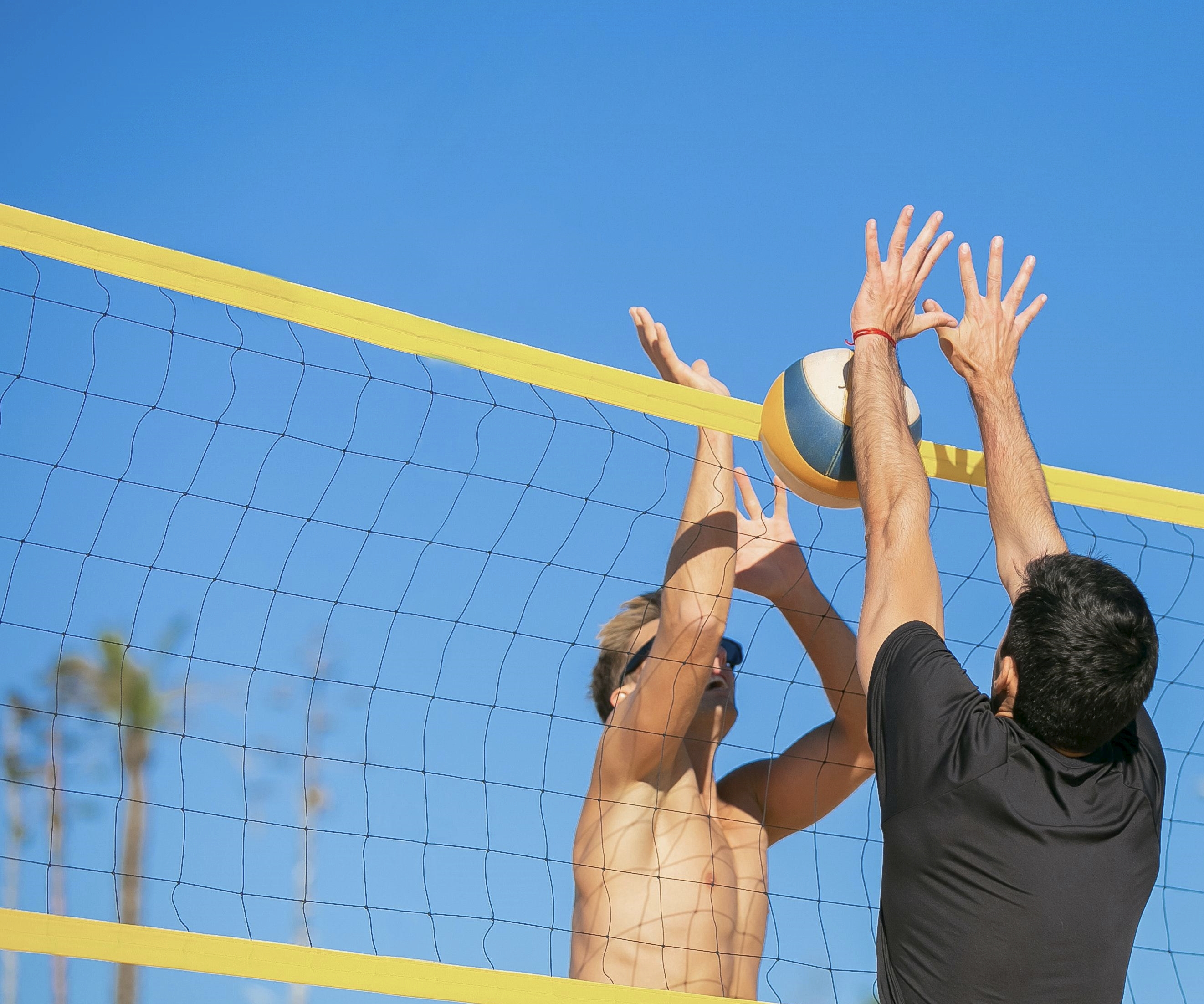}
    \end{minipage}\hfill
    \begin{minipage}[t]{0.73\textwidth}
        \vspace{0pt}
        \small
        \textbf{Query:} \textit{Describe this image.}

        \vspace{4pt}
        \answerbox{red!3}{Base}{The image captures a lively \halluc{beach}
        \grounded{volleyball} game in progress. Two men are playing against
        each other, both reaching for the same \grounded{volleyball} that is in
        the air. They are both wearing sunglasses, and their athletic attire
        suggests they are fully engaged in the game. The \grounded{volleyball}
        is positioned in the middle of the scene, with both players trying to
        make a successful hit. The \grounded{net} is visible in the background,
        and the \halluc{beach} setting adds to the excitement of the game.}

        \vspace{5pt}
        \answerbox{blue!3}{$+$\methodname}{The image captures a dynamic moment
        during a \grounded{volleyball} game. Two men are playing against each
        other, both reaching for the same \grounded{volleyball} that is in the
        air. They are both wearing black shirts and are positioned close to the
        \grounded{net}. The \grounded{volleyball} is located in the center of
        the scene, with both players trying to grab it. The \grounded{net} is
        visible in the background, and the two men are the main focus of the
        image as they compete for the ball.}
    \end{minipage}
    \caption{\textbf{Generative example from AMBER.} Red marks terms from
    the annotated hallucination vocabulary, and blue marks annotated
    ground-truth objects.}
    \label{fig:amber-examples}
\end{figure*}

Figure~\ref{fig:amber-examples} shows a generative example from AMBER, whose
per-image annotations provide both a ground-truth object list and a
hallucination vocabulary. The pattern mirrors the CHAIR examples. The base
model imports a scene-typical beach setting that the annotations flag, while
\methodname{} avoids the flagged term. The candidate example was flagged
mechanically from the annotation vocabulary and then manually verified
against the image before inclusion.

\subsection{Failure Modes of Ablated Components}
\label{sec:failure-modes}

\begin{figure*}[t]
    \centering
    \exampleheader{(a)}{\llavaonefive}
    \vspace{5pt}

    \begin{minipage}[t]{0.24\textwidth}
        \vspace{0pt}
        \centering
        \includegraphics[width=\linewidth]{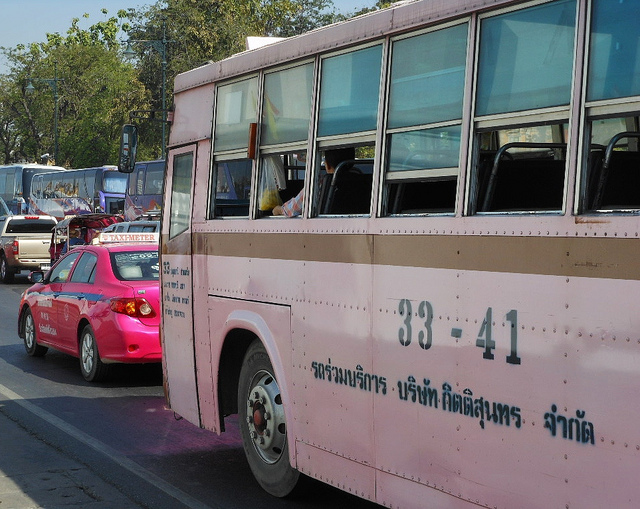}
    \end{minipage}\hfill
    \begin{minipage}[t]{0.73\textwidth}
        \vspace{0pt}
        \small
        \textbf{Unbounded update $\rightarrow$ repetition collapse}

        \vspace{4pt}
        \answerbox{red!3}{$-$Bounding (unbounded update)}{The image depicts a
        busy city scene with multiple buses, cars, a bicycle, motorcycle,
        truck, motorcycle, taxi, [\dots] \halluc{bus}, \halluc{car},
        \halluc{bus}, \halluc{car}, \halluc{bus}, \halluc{car}, \halluc{bus},
        \halluc{car}, \halluc{bus}, \halluc{car}, \halluc{bus}, \halluc{car},
        \halluc{bus}, \halluc{car}, \halluc{bus}, \halluc{car}, \halluc{bus},
        \halluc{car}, \halluc{bus}, \halluc{car}, \halluc{bus}, \halluc{car},
        \halluc{bus}, \halluc{car}, \halluc{bus}, \halluc{car}, \halluc{bus},
        \halluc{car}, \halluc{bus}, \halluc{car}}

        \vspace{5pt}
        \answerbox{blue!3}{$+$\methodname{} (full)}{The image depicts a busy
        city street filled with various vehicles, including a pink
        \grounded{bus}, several \grounded{cars}, and a \grounded{truck}. The
        pink \grounded{bus} is driving down the street, surrounded by other
        vehicles. There are multiple \grounded{people} visible in the scene,
        some of them standing or walking near the vehicles, while others are
        inside the \grounded{bus} or other vehicles.  The street is quite
        congested, with traffic in both directions, including \grounded{cars}
        and \grounded{buses}. The scene captures the bustling atmosphere of a
        city, with \grounded{people} going about their daily routines and
        commuting using different modes of transportation.}
    \end{minipage}

    \vspace{10pt}
    \exampleheader{(b)}{\llavaonefive}
    \vspace{5pt}

    \begin{minipage}[t]{0.24\textwidth}
        \vspace{0pt}
        \centering
        \includegraphics[width=\linewidth]{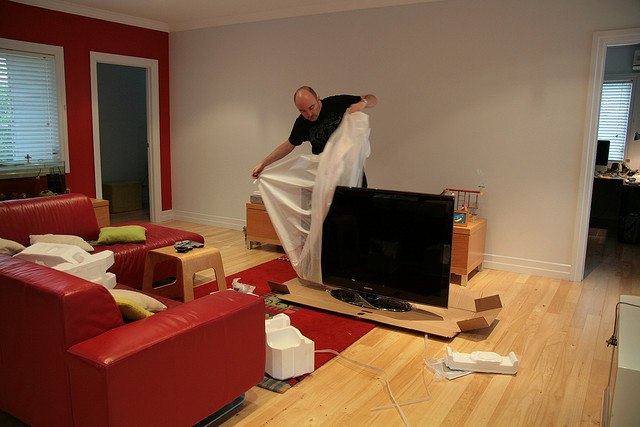}
    \end{minipage}\hfill
    \begin{minipage}[t]{0.73\textwidth}
        \vspace{0pt}
        \small
        \textbf{Global reference $\rightarrow$ misdirected suppression}

        \vspace{4pt}
        \answerbox{red!3}{$-$Token reference (global reference)}{The image
        shows a \grounded{man} in a black shirt standing in a living room,
        unboxing a large flat-screen \grounded{TV}. He is carefully removing
        the plastic covering from the \grounded{TV}, which is placed on a
        wooden \halluc{table}. The room is furnished with a \grounded{couch}
        and a \halluc{chair}, and there is a \halluc{remote} control on the
        \grounded{couch}.  In the background, there is a \halluc{dining table}
        with a \halluc{bottle} on it. The living room also features a
        \halluc{potted plant}, a \halluc{vase}, and a \halluc{book}, adding to
        the cozy atmosphere of the space.}

        \vspace{5pt}
        \answerbox{blue!3}{$+$\methodname{} (full)}{The image shows a
        \grounded{man} standing in a living room, carefully wrapping a large
        flat-screen \grounded{TV} in plastic. The \grounded{TV} is placed on a
        wooden stand, and the \grounded{man} is using plastic to protect it
        from dust and potential damage during transportation.  The living room
        is furnished with a \grounded{couch} and a \halluc{chair}, both
        located towards the left side of the room. There is also a
        \halluc{remote} control placed on the \grounded{couch}, likely for
        operating the \grounded{TV} or other electronic devices. The overall
        atmosphere suggests that the \grounded{man} is taking care to ensure
        the \grounded{TV}'s safe and secure transportation to its new
        location.}
    \end{minipage}
    \caption{\textbf{Failure modes of ablated components on
    \llavaonefive{}.} Arm definitions are in
    Table~\ref{tab:ablation-definitions}. In (a), the arm caption is
    abridged, and the omitted middle, marked [\dots], repeats only the two
    tokens ``bus'' and ``car''. Red in (a) marks the repeated tokens in the
    shown tail. In (b), red marks CHAIR-hallucinated objects, and blue marks
    ground-truth objects.}
    \label{fig:failure-modes}
\end{figure*}

Figure~\ref{fig:failure-modes} illustrates, on actual generations, the two
characteristic failure modes behind the component ablation in
Table~\ref{tab:signal-ablation}. Removing the bound (\emph{Unbounded update})
turns suppression into unbounded logit displacement. Once the evidence-poor
continuation is displaced far enough, the decoder falls into a two-token loop
and repeats the same objects until the length limit, with a maximum
single-object repetition of 122. This collapse inflates caption length and
drives the mention-level hallucination rate far above the base model. Replacing the token-specific
reference $b(v)$ with a single global reference (\emph{Global reference})
misallocates suppression across candidates. On the example shown, it derails
an otherwise clean continuation into five object classes that neither the
base model nor the full method produces (book, bottle, dining table, potted
plant, and vase). The full method keeps the base model's two flagged
objects (chair and remote) and adds none. Both examples are the top-ranked
cases under a preregistered mechanical ordering rather than hand-picked
selections. The ordering ranks degeneration cases by their maximum
single-object repetition and misallocation cases by the number of newly
introduced object classes, breaking ties by image ID. The corresponding
aggregate diagnostics appear in the ablation diagnostics tables.

\let\grounded\relax
\let\halluc\relax
\let\exampleheader\relax
\let\answerbox\relax

\section{Use of AI Assistants}
\label{sec:ai-assistance}

In preparing this manuscript, we used AI-based writing assistants in a
limited capacity to suggest alternative phrasings, correct grammatical
errors, and improve readability. The authors reviewed and approved all such
suggestions. AI assistants were not used to formulate the research ideas,
design the method or experiments, implement the system, or generate or
verify citations, factual claims, or dataset statistics.

Separately from writing assistance, the official protocols of MMHal-Bench
and MM-Vet specify an LLM-based automatic evaluator. Following these
protocols, we score both benchmarks with the frozen
\texttt{gpt-4.1-mini-2025-04-14} judge under identical settings for every
compared method (Appendix~\ref{sec:evaluation-prompts}). This judge is a
component of the evaluation pipeline defined by the benchmarks and is
distinct from the writing assistance disclosed above. All other reported
quantities are computed by deterministic scorers that involve no
generative model.

All technical content, implementation details, analyses, and reported
findings were produced and verified by the authors, who take full
responsibility for the manuscript.

\end{document}